\documentclass[a4paper,fleqn]{cas-sc}

\usepackage[numbers,sort&compress]{natbib}

\usepackage{amsthm}
\usepackage{algorithm}
\usepackage{algorithmic}
\usepackage{subcaption}
\usepackage{tabularx}
\usepackage{makecell}
\usepackage{placeins}
\usepackage{float}
\usepackage{textcomp}
\usepackage{url}
\usepackage{orcidlink}
\usepackage{listings}
\usepackage{tcolorbox}
\newcommand{\tcheck}[1]{#1} 
\usepackage{tikz}
\usetikzlibrary{shapes.geometric, arrows.meta, positioning, calc, fit}
\usepackage{pgfplots}
\pgfplotsset{compat=1.18}

\definecolor{pVerde}{HTML}{D6336C}
\definecolor{pVerdeOsc}{HTML}{8E1F4F}
\definecolor{pSalvia}{HTML}{FDC5F5}
\definecolor{pAzul}{HTML}{8093F1}
\definecolor{pCiruela}{HTML}{B388EB}
\definecolor{pGris}{HTML}{2F2A2B}

\lstdefinestyle{python}{
    language=Python,
    basicstyle=\ttfamily\small,
    keywordstyle=\color{blue}\bfseries,
    commentstyle=\color{gray}\itshape,
    stringstyle=\color{orange},
    numberstyle=\tiny\color{gray},
    numbers=left, numbersep=5pt, frame=single, framesep=3pt,
    backgroundcolor=\color{gray!5},
    breaklines=true, breakatwhitespace=true, tabsize=4,
    showstringspaces=false,
    morekeywords={ctx, self, True, False, None},
}

\theoremstyle{definition}

\begin{document}

\let\WriteBookmarks\relax

\renewcommand{\topfraction}{.95}
\renewcommand{\bottomfraction}{.85}
\renewcommand{\textfraction}{.05}
\renewcommand{\floatpagefraction}{.75}
\setcounter{topnumber}{5}
\setcounter{bottomnumber}{5}
\setcounter{totalnumber}{10}

\shorttitle{Transferable knowledge graphs with executable learned operators for algorithm design}
\shortauthors{C. Chac\'on Sartori et al.}

\title[mode=title]{Transferable knowledge graphs with executable learned operators for algorithm design}

\author[1]{Camilo Chac\'on Sartori}[orcid=0000-0002-8543-9893]
\cormark[1]
\credit{Conceptualization, Methodology, Software, Writing -- original draft}
\ead{camilo.chacon@icn2.cat}
\affiliation[1]{organization={Catalan Institute of Nanoscience and Nanotechnology (ICN2), CSIC and BIST},
            addressline={Campus UAB},
            city={Bellaterra},
            postcode={08193},
            state={Barcelona},
            country={Spain}}

\author[1]{Jos\'e H. Garc\'ia}[orcid=0000-0002-5752-4759]
\credit{Methodology, Supervision}
\ead{josehugo.garcia@icn2.cat}

\author[1]{Andrei Voicu Tomut}[orcid=0000-0001-8288-4054]
\credit{Methodology, Supervision}
\ead{andrei.tomut@icn2.cat}

\author[2]{Christian Blum}[orcid=0000-0002-1736-3559]
\credit{Supervision, Writing -- review and editing}
\ead{christian.blum@iiia.csic.es}
\affiliation[2]{organization={Artificial Intelligence Research Institute (IIIA-CSIC)},
            city={Bellaterra},
            country={Spain}}

\cortext[1]{Corresponding author}

\begin{abstract}
Procedural knowledge in algorithm design is embedded in source code and rebuilt for each new domain. We introduce Generative Executable Algorithm Knowledge Graphs (GEAKG), a representation in which this knowledge is stored as a generative, executable, transferable graph: typed nodes hold validated operators, edges encode admissible compositions, and learned edge weights record effective sequences. The same engine instantiates the structure across domains by changing only a role ontology (RoleSchema) and a binding. We study GEAKG as a representation mechanism rather than a state-of-the-art optimizer, asking what transfers and when. Layer ablations localize transfer by granularity: within a neural-architecture-search family the learned snapshot transfers across 70 dataset pairs---its weights stay correlated across datasets and one frozen snapshot remains competitive with Regularized Evolution at zero deployment-token cost; across combinatorial domains only the ontology-constrained executable structure transfers, not the learned weights. That structure pays off where target-side search is expensive---a Traveling Salesman snapshot beats an equally untuned from-scratch search on large scheduling instances even at one-fifth its budget---but does not improve on an effective local search where one is cheap, as in assignment and linear ordering. Executable procedural knowledge can thus be acquired offline, compacted, inspected, and reused without runtime language-model calls.
\end{abstract}

\begin{highlights}
\item Procedural knowledge is represented as executable, typed knowledge graphs
\item One engine spans neural architecture search and optimization at zero runtime language-model tokens
\item What transfers depends on granularity: learned weights within a family, executable structure across domains
\item Where transfer pays is set by target-side search cost: it helps on large-scale scheduling, not on assignment or linear ordering
\end{highlights}


\maketitle

\printcredits

\section{Introduction}
\label{sec:introduction}
Knowledge graphs organize structured knowledge through entities, relations, and schema constraints~\cite{hogan2021knowledge}. Algorithm design depends on another kind of knowledge: the procedural choices that determine which operation should be applied, in what order, and under which state. That knowledge is usually embedded in source code, experiment logs, or tacit design practice. When the domain changes, the procedure is commonly rebuilt, tuned, or evolved again.

We introduce \textit{Generative Executable Algorithm Knowledge Graphs} (GEAKG), a knowledge-representation paradigm for procedural algorithm-design knowledge. A GEAKG is a typed directed graph in which nodes correspond to abstract operator roles, node payloads are validated executable operators, and edge weights encode learned evidence about productive operator transitions. The representation has three properties. It is \textit{generative}: topology and operators can be synthesized by a language model under schema constraints. It is \textit{executable}: traversal invokes runnable code. It is \textit{transferable}: a learned snapshot can move to a new compatible domain by changing the binding while keeping topology, operators, and transition weights fixed.

The key separation is between a domain-agnostic engine and a domain-specific ontology. The engine implements three layers. L0 is a MetaGraph of valid role transitions. L1 is a pool of executable operators bound to those roles. L2 is learned procedural knowledge, represented by edge weights and symbolic rules refined through Ant Colony Optimization (ACO). A RoleSchema defines the role vocabulary, categories, entry points, revisitability constraints, and admissible transitions. Once a RoleSchema and a domain binding are supplied, the same construction, learning, snapshotting, and symbolic-execution machinery applies. GEAKG is therefore proposed as a way to acquire, store, inspect, and redeploy procedural knowledge, not as a specialized solver or an AutoML system.

Zero-shot transfer has a precise meaning here. A target-domain deployment uses a fixed snapshot consisting of L0, L1, and L2. There is no target-domain edge weight retraining, no operator resynthesis, and no language-model call at runtime. The only target-specific element is a schema-compatible binding that provides evaluation, validity checking, copying, and related domain operations. RoleSchema design is a one-time modeling step and is reported separately from deployment cost.

\begin{figure}[pos=!htbp]
\centering
\begin{subfigure}[t]{0.45\textwidth}
\centering
\includegraphics[width=\textwidth,height=0.36\textheight,keepaspectratio]{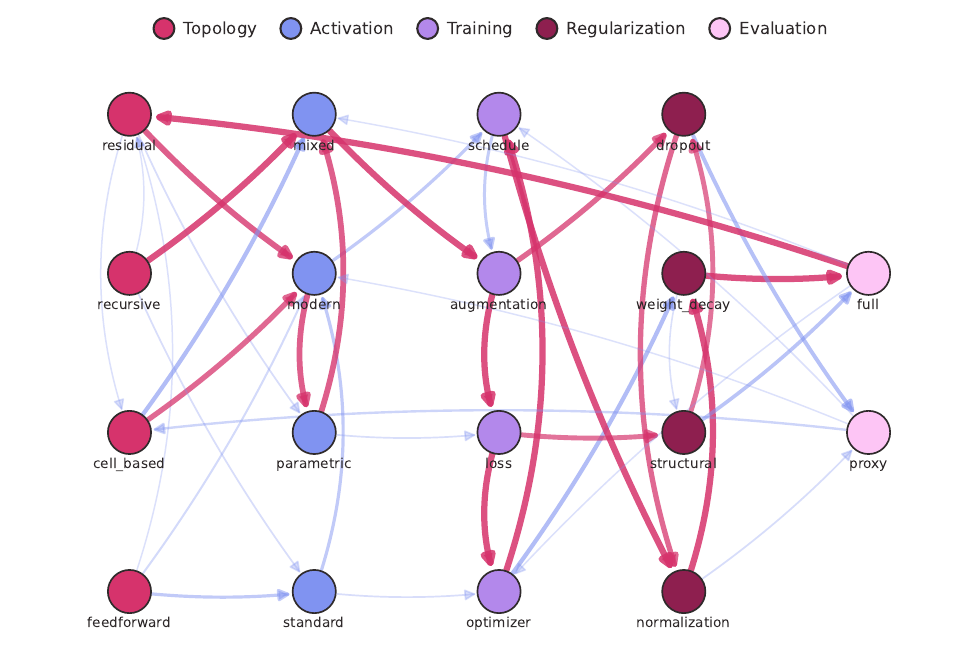}
\caption{NAS RoleSchema instantiation.}
\label{fig:metagraph_nas}
\end{subfigure}
\hfill
\begin{subfigure}[t]{0.45\textwidth}
\centering
\includegraphics[width=\textwidth,height=0.36\textheight,keepaspectratio]{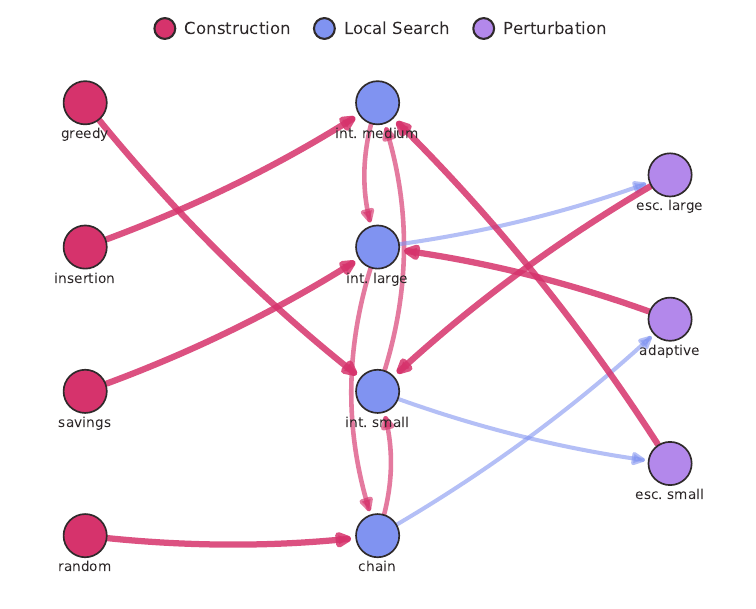}
\caption{Optimization RoleSchema instantiation.}
\label{fig:metagraph_opt}
\end{subfigure}
\caption{Two MetaGraphs produced by the same GEAKG engine from different RoleSchemas.}
\label{fig:dual_case_study}
\end{figure}

Figure~\ref{fig:geakg_pipeline} summarizes the pipeline. Offline, a RoleSchema conditions topology and operator generation; generated operators are validated; ACO refines transition weights over execution traces; and the resulting snapshot stores the reusable procedural artifact. Online, the Symbolic Executor traverses the snapshot and applies the bound operators in a target domain. Figures~\ref{fig:dual_case_study}a and~\ref{fig:dual_case_study}b show two instantiations of the same engine: neural architecture search (NAS), where roles encode architecture-design decisions, and combinatorial optimization, where roles encode construction, local search, and perturbation.

\begin{figure}[pos=!htbp]
\centering
\begin{tikzpicture}[
box/.style={rectangle,draw,thick,rounded corners=3pt,minimum height=0.9cm,align=center,font=\footnotesize},
arrow/.style={->,>=stealth,thick}]
\node[box,fill=gray!15] (s) at (-5,2) {RoleSchema\\ontology};
\node[box,fill=pAzul!20] (l0) at (-2,2) {L0 MetaGraph\\topology};
\node[box,fill=pVerde!20] (l1) at (1,2) {L1 Operators\\validated code};
\node[box,fill=pCiruela!20] (l2) at (4,2) {L2 Learning\\edge weights};
\node[box,fill=pVerdeOsc!20] (snap) at (1,0) {GEAKG snapshot};
\node[box,fill=pAzul!15] (ex) at (-1,-2) {Symbolic Executor};
\node[box,fill=pVerde!15] (ctx) at (3,-2) {Domain binding};
\draw[arrow](s)--(l0);\draw[arrow](l0)--(l1);\draw[arrow](l1)--(l2);\draw[arrow](l2)--(snap);\draw[arrow](snap)--(ex);\draw[arrow,<->](ex)--(ctx);
\end{tikzpicture}
\caption{End-to-end GEAKG pipeline. Offline generation and ACO learning produce a snapshot; online symbolic execution applies it through a domain binding with zero runtime language-model tokens.}
\label{fig:geakg_pipeline}
\end{figure}

The empirical goal is deliberately scoped: we ask whether algorithmic know-how can be captured as an executable graph and reused at negligible marginal cost, rather than whether the resulting solver outperforms specialized state-of-the-art methods. We evaluate three research questions. RQ1 asks whether one engine can support different algorithm-design domains without framework-level changes. RQ2 asks whether procedural snapshots transfer across datasets and compatible optimization domains. RQ3 asks whether the transferred artifact persists as a compact, inspectable, zero-token deployment unit.

The central finding has two parts. First, transfer has a clear granularity. Across domains, the reusable content is the ontology-constrained executable structure: the L0 topology and L1 operators. The L2 learned transition weights refine performance within the source domain, but they do not carry the cross-domain gain. Second, this structure pays off when target-side search is expensive. It beats an equally untuned from-scratch search on large job-shop instances, even when the baseline is given five times the budget. It is matched where cheap local search is already effective, as on quadratic assignment and linear ordering.

The evidence is organized around this decomposition. The supporting contributions are a formal GEAKG representation; two instantiations using the same engine; NAS transfer across 70 dataset pairs; cross-domain optimization transfer under strict precedence-aware evaluation, mapping where transfer pays off (scheduling) and where it meets a boundary (assignment, linear ordering); design ablations isolating each layer (L0 ontology, L1 operators, L2 edge weights); and artifact analysis through structure, spectral stability, edge weight maps, and rule quality. \tcheck{Table~\ref{tab:claim_evidence} maps each central claim to the controlled comparison that tests it and to its outcome.}

\begin{table}[pos=!htbp]
\centering
\caption{Claim-to-evidence map: each central claim, the controlled comparison that tests it, and the outcome. ``$\approx$'' denotes statistically indistinguishable.}
\label{tab:claim_evidence}
\small
\begin{tabular}{@{}p{0.29\columnwidth}p{0.40\columnwidth}p{0.23\columnwidth}@{}}
\toprule
Claim & Test (vs.\ comparator) & Result \\
\midrule
One engine, multiple domains & same pipeline on NAS and optimization & runs unchanged on NAS + 3 optimization domains \\
Within-family: weights transfer & NAS cross-dataset vs.\ random ordering & 62/70 significant; $\bar{r}=0.91$ \\
Cross-domain: only structure transfers & L2 learned vs.\ uniform weights (JSSP, QAP, LOP) & learned $\approx$ uniform ($p=0.71,\,0.64$) \\
Each layer is load-bearing & L0/L1/L2 ablations & L0 lowers gap; no-op L1 collapses; L2 in-domain only \\
Transfer pays where search is expensive & JSSP vs.\ untuned ILS at $1\times$ and $5\times$ budget & GEAKG wins even at ILS $5\times$ \\
Boundary where search is cheap & QAP/LOP vs.\ ILS at one-fifth budget & ILS at $\tfrac{1}{5}$ budget still wins \\
Zero marginal deployment cost & vs.\ Regularized Evolution, Bayesian optimization & 0 language-model tokens; seconds \\
\bottomrule
\end{tabular}
\end{table}

\section{Related work and positioning}
\label{sec:related_work}
Existing KG paradigms are predominantly declarative. Traditional KGs, including Freebase, Wikidata, and ConceptNet~\cite{speer2017conceptnet}, store entity--relation triples and answer ``what is'' queries. KG embeddings such as TransE~\cite{bordes2013translating} learn dense vectors for link prediction over declarative graphs, and rule-mining systems such as AMIE~\cite{galarraga2013amie} mine logical rules from graph structure. GEAKG instead learns scalar edge weights over a procedural graph whose edges represent operator transitions, and its rules are mined from execution traces.

Executable KG systems such as ExeKGLib~\cite{10.1145/3511808.3557195} move closer to computation, but their nodes are pipeline specifications or code-derived records rather than directly runnable algorithmic procedures. GEAKG nodes store runnable operators, while topology and operators are generated under the RoleSchema ontology. Process-mining KGs~\cite{vanderaalst2016process,bachhofner2022automated} are descriptive: they summarize observed behavior. GEAKG is generative in the procedural sense: learned weights guide new procedure construction during traversal. Workflow-provenance models such as PROV~\cite{moreau2013prov} and ProvONE~\cite{cuevasvicenttin2016provone} capture retrospective lineage, but they do not learn edge-level decision policies from optimization outcomes. GEAKG's edge weights encode which sequences have been effective for prospective transition selection.

Following Hogan et al.'s taxonomy~\cite{hogan2021knowledge}, GEAKG is a domain-specific KG whose schema is the RoleSchema ontology, whose nodes store executable artifacts, and whose edges are refined by ACO, a form of KG refinement~\cite{paulheim2017knowledge}. The graph prescribes procedures rather than describing only facts or provenance. The same distinction separates GEAKG from code-evolution systems such as LLaMEA~\cite{10752628}, FunSearch~\cite{Romera-Paredes2024}, EoH~\cite{10.5555/3692070.3693374}, ReEvo~\cite{ye2024reevo}, and irace-evo, which embeds code evolution in automatic algorithm configuration~\cite{sartori2025iraceevoautomaticalgorithmconfiguration}. Those systems keep knowledge implicit in generated programs. GEAKG instead makes operators components inside a persistent, inspectable graph.

GEAKG also differs from graph-based optimization methods. In those methods, the graph is usually the problem instance or a learned policy network being optimized. Here, the graph is the transferable procedural artifact, and its nodes are executable algorithm-design operators rather than problem variables or network parameters. Ant Colony Optimization supplies one component of GEAKG: the L2 rule for learning transition weights. Classical ACO rebuilds a graph over problem-specific solution components, such as cities or edges, and learns a scalar over those components. GEAKG integrates three layers that classical ACO does not have: an ontology of abstract, domain-agnostic operator roles (L0), a pool of language-model-generated executable operators bound to those roles (L1), and learned transition weights (L2). These layers are serialized together as an inspectable artifact that can be rebound to new domains with no further generation. The contribution is the integrated, transferable representation.

\section{Results}
\label{sec:results}
\subsection{One engine, two domains}
NAS and combinatorial optimization use the same GEAKG construction pipeline, snapshot format, ACO learning loop, and Symbolic Executor. The domain-specific parts are limited to the RoleSchema and the binding used to evaluate candidate objects. In NAS, roles encode topology choice, activation selection, training policy, regularization, and evaluation. In optimization, roles encode construction, intensification, and perturbation operators for permutation-based search. This shared infrastructure supports the representation claim: procedural knowledge can be stored in a graph whose execution engine is independent of the target domain once the role ontology and evaluator are supplied.

In both case studies, L0 supplies the typed transition envelope, L1 supplies validated executable operations, and L2 supplies learned edge preferences. The snapshot is then applied by the same symbolic traversal routine. This separation is central for transfer. If a target domain exposes the required context interface and shares the same abstract role vocabulary, the learned transition structure can be reused without target-domain generation.

The two domains also exercise different notions of executability. In NAS, execution means applying graph-guided architecture transformations and evaluating the resulting architecture through a tabular benchmark. The graph therefore acts at the level of search-space navigation: traversal selects the next design role by sampling the learned transition distribution and applies the bound operator to produce an architecture specification. In optimization, execution is closer to direct solution manipulation: operators construct, perturb, and locally improve permutations that are evaluated immediately by a domain context. The same GEAKG abstraction covers both cases because the executor only requires role-bound operators and a scalar feedback signal. This is the empirical basis for treating GEAKG as a procedural-knowledge representation rather than a domain-specific algorithm.

\subsection{Neural architecture search}
The NAS case study evaluates whether a GEAKG snapshot provides a reusable search policy over architecture-design decisions. The same NAS RoleSchema contains 18 roles in 5 categories and is used for both graph neural network and convolutional cell search. NAS-Bench-Graph provides 26,206 graph neural architectures across graph datasets~\cite{qin2022nasbenchgraph}; NAS-Bench-201 provides 15,625 cell architectures across CIFAR-10, CIFAR-100, and ImageNet16-120~\cite{dong2020nasbench201}. Each method receives the same architecture-evaluation budget, and the Symbolic Executor uses a frozen snapshot at deployment.

\begin{figure}[pos=!htbp]
\centering
\includegraphics[width=0.75\columnwidth]{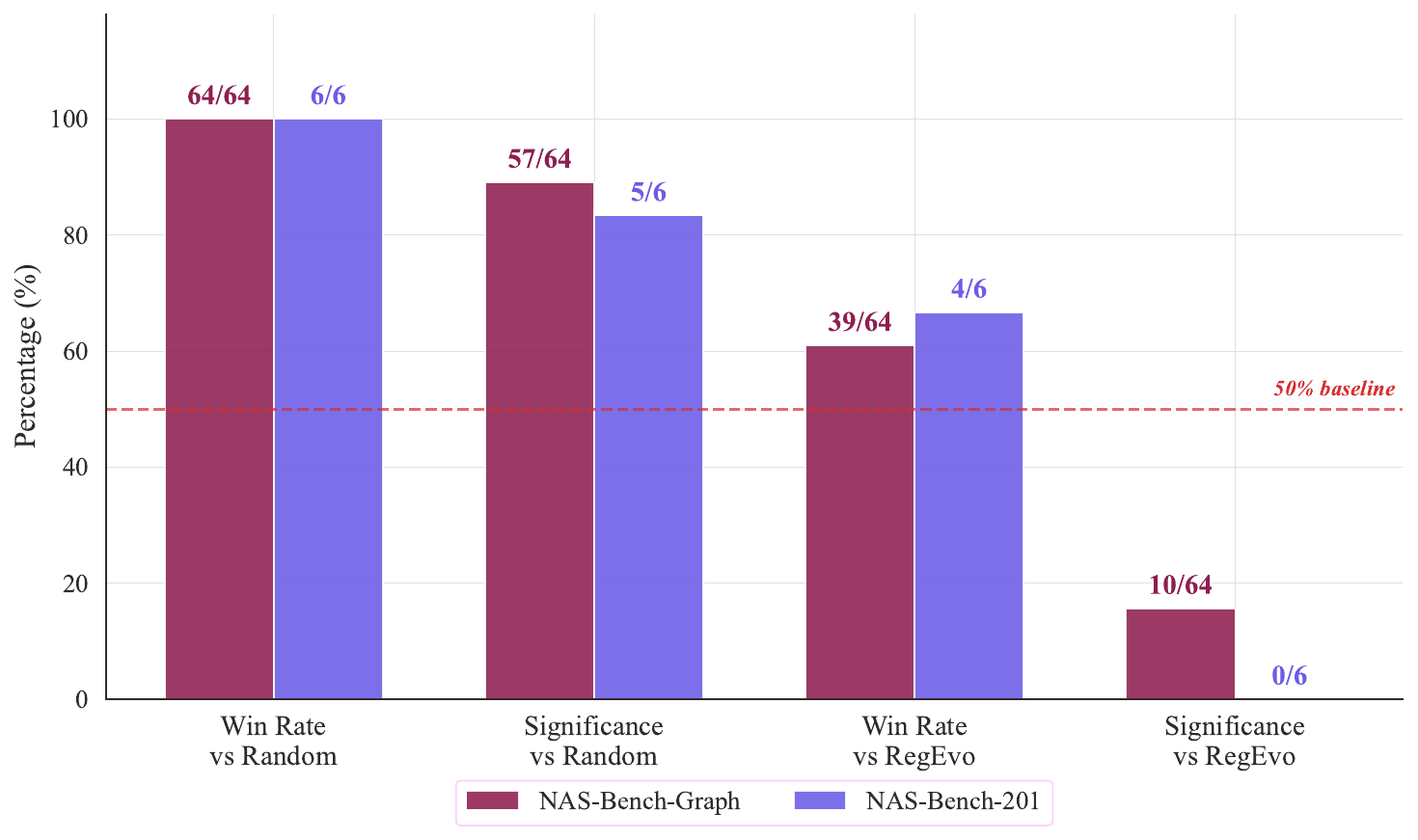}
\caption{Aggregate comparison across the 70 NAS transfer pairs. The Symbolic Executor improves over Random Search on every pair and remains competitive with Regularized Evolution while using a frozen GEAKG snapshot at deployment.}
\label{fig:nas_aggregate}
\end{figure}

Across both benchmarks, the Symbolic Executor improves over Random Search on all 70 transfer pairs (Figure~\ref{fig:nas_aggregate}). The comparison to Random Search is best interpreted as a sequence ablation: the operator pool is shared, while the ordering policy changes from random selection to learned edge-weight-guided traversal. The per-benchmark results sharpen this picture. On NAS-Bench-Graph, the 64 transfer configurations consist of graph neural architecture transfers over the 26,206-architecture search space. GEAKG improves over Random Search in 64/64 pairs, with 57/64 statistically significant at $p<0.05$ and a median accuracy delta of +1.26 percentage points (Supplementary Fig.~S2). Against Regularized Evolution~\cite{real2019regularized} on the same 64 NAS-Bench-Graph pairs, the GEAKG snapshot wins 39 pairs in mean accuracy and reaches statistical significance in 10 pairs, with a median accuracy delta of +0.13 percentage points. This is a competitive rather than dominant result. Regularized Evolution performs target-side evolutionary search, while GEAKG deploys a frozen procedural artifact with zero language-model tokens and negligible marginal runtime.

NAS-Bench-201 provides a smaller but complementary convolutional-cell test. Its 15,625-architecture search space has a compressed accuracy range, so absolute differences between methods are smaller. Across the 6 NAS-Bench-201 transfer pairs, GEAKG improves over Random Search in 6/6 pairs, with 5/6 statistically significant and a mean accuracy delta of +0.84 percentage points (Supplementary Fig.~S3). Against Regularized Evolution, GEAKG wins 4/6 pairs in mean accuracy, none reach statistical significance, and the mean delta is +0.06 percentage points. These results support the same interpretation as NAS-Bench-Graph but with a narrower margin: the transferred graph supplies a reusable search prior, while the strongest evolutionary baseline remains close under a matched evaluation budget.

The two NAS benchmarks also test framework reuse across architecture families. NAS-Bench-Graph uses 6-node directed acyclic graphs with 9 operations, whereas NAS-Bench-201 uses 4-node cells with 5 operations. GEAKG keeps the same 18-role NAS RoleSchema, the same Symbolic Executor implementation, and the same L1 pool of 28 operators synthesized with gpt-5.2. The MetaGraph differs in realized edge count, with 42 edges in the graph case and 32 in the cell case, but the engine-level code path is shared. The domain-specific elements are the architecture representation, the base operators, and the tabular evaluator. This is the most direct evidence for RQ1 in the NAS case: the same procedural roles can organize GNN and CNN architecture search even though the underlying candidate encodings differ.

Because both NAS benchmarks are tabular, we also evaluate the fidelity of the tabular signal used by GEAKG. Table~\ref{tab:nas_fidelity} reports rank correlations for NAS-Bench-201/NATS-Bench over 1,500 architectures per dataset. The 200-epoch accuracy that GEAKG optimizes is highly reproducible across independent training seeds, with Spearman correlations from 0.982 to 0.988. The lower-fidelity 12-epoch proxy, which GEAKG does not use, correlates at 0.812 to 0.875 with the 200-epoch value. These numbers support the use of the tabular 200-epoch signal as a high-fidelity benchmark target while leaving full retraining under a new pipeline for future work.

\begin{table}[pos=!htbp]
\centering
\caption{Surrogate fidelity of NAS-Bench-201/NATS-Bench tabular accuracies over 1,500 architectures per dataset.}
\label{tab:nas_fidelity}
\small
\begin{tabular}{@{}lcccc@{}}
\toprule
Dataset & \makecell{Cross-seed\\$\rho$ (200ep)} & \makecell{12ep$\to$200ep\\$\rho$} & \makecell{Seed noise\\(pp)} & \makecell{Top-10\\overlap} \\
\midrule
CIFAR-10 & 0.985 & 0.875 & 0.17 & 0.40 \\
CIFAR-100 & 0.982 & 0.829 & 0.33 & 0.40 \\
ImageNet16-120 & 0.988 & 0.812 & 0.37 & 0.50 \\
\bottomrule
\end{tabular}
\end{table}

The NAS evidence supports two conclusions. First, learned transition weights matter within this source family: replacing them with random ordering removes the advantage across all transfer pairs. Second, the practical value is amortization. L1 generation is a one-time offline language-model cost; ACO learning uses no language-model calls, and deployment uses none. The snapshot is a compact procedural prior that can be applied repeatedly across datasets.

The fidelity check also applies to GEAKG's selected architectures. On the CIFAR-10 source, the Symbolic Executor reaches 94.34\%, within 0.03 percentage points of the global real-trained optimum of 94.37\%. Under zero-shot transfer to CIFAR-100, it attains $71.13\pm0.11\%$ compared with Regularized Evolution's 70.43\% at a matched 50-evaluation budget. The run-to-run standard deviation of 0.11 percentage points is smaller than the benchmark's own CIFAR-100 seed noise of 0.33 percentage points. The caveat is that the identity of the single best architecture is seed-sensitive: only 40--50\% of the top-10 architectures recur across seeds in Table~\ref{tab:nas_fidelity}. The appropriate claim is therefore rank-faithful top-tier selection, rather than exact recovery of a unique optimum.

The Random Search comparison is intentionally a low but important bar. Random Search samples from the same benchmark spaces without the learned transition weights. GEAKG's 70/70 mean-accuracy improvement over that baseline, with 62/70 significant at $p<0.05$, shows that the stored transition weights affect search behavior in a consistent direction. The stronger RegEvo comparison calibrates the magnitude of the result: on NAS-Bench-Graph, the median +0.13 percentage-point delta is small, and only 10/64 pairs reach significance. The result is the expected signature of a reusable prior that competes with a target-side optimizer while full per-target search remains useful in some settings.

These accuracy results come at negligible deployment cost: a frozen snapshot completes 500 evaluations in seconds and matches unlimited Bayesian Optimization accuracy. At the same time, that baseline needs roughly 22 minutes of per-target search. Full transfer-layout, timing, surrogate-fidelity, edge-weight stability, and variance detail is reported in Methods (NAS transfer protocol, deployment cost, and stability).

\subsection{Optimization transfer}
This case study is a boundary experiment. A single snapshot learned on the Traveling Salesman Problem is redeployed, unchanged, on three permutation-encoded targets: job-shop scheduling, quadratic assignment, and linear ordering. The setup asks which layer carries across domains and when reuse pays. TSP is the source because it provides a stable setting for offline learning and a well-studied objective. On small TSP instances, GEAKG matches strong code-evolution performance from LLaMEA; on larger instances, it remains robust when LLaMEA fails to return valid tours within the token budget. A hybrid configuration that embeds one evolved operator wins 5/7 TSP instances under a matched 50k-token budget (Supplementary Tables~S1 and~S2). GEAKG also returns feasible tours on all seven (7/7) TSP instances across the gpt-5.2, gpt-4o-mini, and fully local Qwen2.5-14B settings, whereas standalone LLaMEA returns a valid tour on only 1/7 instances in the local 10k-token configuration. With L0 and L2 fixed, the single embedded LLaMEA operator drives the largest gap reductions over standalone LLaMEA, 65\% on pr226 and 54\% on pcb442. These source results make the snapshot useful enough to test as a reusable procedural artifact. They also separate operator quality from transfer: better L1 code improves the source-domain solver, while the cross-domain experiment below asks whether the stored graph structure remains useful after the objective changes.

The transfer test keeps L0, L1, and L2 fixed; target domains supply only evaluation and validity logic. No target-domain edge weight retraining, operator resynthesis, or runtime language-model call is used. JSSP and QAP share the permutation binding with TSP but differ in domain semantics and objective structure~\cite{garey1976complexity,koopmans1957assignment}. The contrast with NAS is important. NAS transfers across datasets within an architecture-search family, so the architecture vocabulary and benchmark interface remain aligned. Optimization transfer changes the objective itself: a TSP tour, a JSSP priority permutation, and a QAP assignment permutation are scored by different functions and constraints. The common substrate is representation-level binding to permutation operations such as swaps, insertions, reversals, and perturbations. This makes the optimization study a stricter test of portability than the NAS study. The experiment therefore tests whether the stored graph structure remains executable when the domain semantics change, and whether the L2 learned transition weights still add value after that change.

Under precedence-aware JSSP evaluation, GEAKG wins 10/14 instances against ILS across eight seeds at 30 seconds per instance. These runs use a makespan evaluation that enforces the job-shop precedence constraints (an operation cannot start before its job predecessor finishes). Neither method is tuned to the target domain: GEAKG deploys a frozen TSP snapshot unchanged, and ILS runs with its default parameters rather than a scheduling-specific configuration. Which method wins depends on instance size. ILS wins the sub-100-operation instances ft06, la01, and la06. At exactly 100 operations the outcome is mixed: GEAKG wins ft10, la16, and abz5, while ILS wins ft20. From 150 operations upward, GEAKG wins la21, la26, la31, la36, la40, abz7, and abz9. The crossover should therefore be read as a trend rather than a threshold rule. The individual large-instance margins reach 22 to 48 percent across the seven instances from 150 operations upward (eight-seed means; per-instance makespans in Supplementary Table~S3). Figure~\ref{fig:transfer_crossover} summarizes this size pattern. A three-seed budget-scaling check on abz7, abz9, and la31 gives the same direction: on the two instances with known optima, GEAKG reaches a $63\%$ gap to optimum while from-scratch ILS reaches $231$--$240\%$ at the same 30-second budget and remains at $150$--$175\%$ when given five times the budget (654--956 iterations rather than 149--231); on la31 the transferred snapshot's makespan is $43\%$ below from-scratch ILS at the same budget and $29\%$ below at five times the budget. The large-instance result is therefore not explained by a shortage of ILS iterations; it reflects a setting where the transferred executable structure remains useful under the fixed wall-clock budget.

\begin{figure}[pos=!htbp]
\centering
\includegraphics[width=0.72\columnwidth]{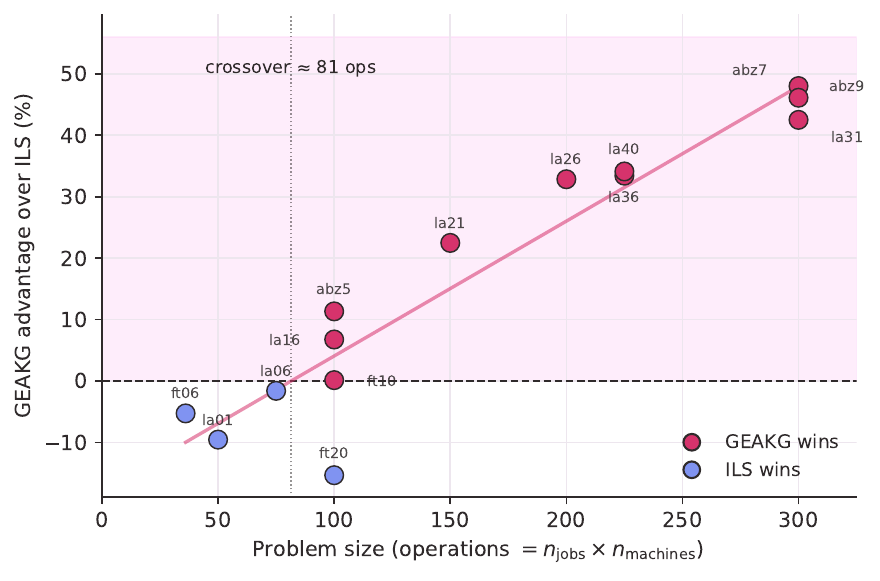}
\caption{Size-dependent zero-shot transfer from TSP to JSSP under precedence-aware evaluation. Each point is one JSSP instance; the vertical axis is GEAKG's makespan advantage over ILS, $(\mathrm{ILS}-\mathrm{GEAKG})/\mathrm{ILS}$. Eight-seed means, 30 seconds per instance; GEAKG wins 10/14 overall.}
\label{fig:transfer_crossover}
\end{figure}

QAP and linear ordering define the boundary. QAP is structurally different from routing and scheduling because each assignment is scored by summed flow times distance over facility pairs; a permutation move therefore changes cost through a quadratic objective. The same TSP snapshot executes through the permutation binding over eight QAPLIB instances from the nug, chr, and tai families at five seeds each (Supplementary Table~S4; an earlier reference evaluation over a wider set of QAP instances is retained for completeness as Supplementary Table~S5). The learned transition weights are indistinguishable from uniform weights (lower mean gap on four of eight instances, paired Wilcoxon $p=0.64$), and a from-scratch ILS is stronger on every QAP instance tested. This matters for interpreting the JSSP result: the same untuned ILS is effective on assignment-style landscapes, so its weaker scheduling performance is not a generic baseline failure. On seven permutation linear-ordering instances ($n=44$--$60$, three seeds; reduced-size variants derived from LOLIB matrices, provided in the Supplementary Data), the transferred snapshot executes, and the learned weighting narrowly outperforms uniform weighting on 13 of 21 runs, but a from-scratch swap-based ILS reaches a higher objective on every instance. The LOP result adds a second assignment-like setting in which the executable structure ports cleanly but the target-side local search remains stronger. Together, these targets preserve the full decomposition: across domains, the reusable content is the ontology-constrained executable structure, not the L2 learned transition weights. The same results give the budget-invariant boundary. Transfer beats untuned from-scratch search where target-side search is expensive, as in large JSSP even at five times the budget. It is matched where cheap local search exists, as in QAP and linear ordering, where one-fifth the budget still beats the transferred snapshot on every instance. This keeps the claim operational: transfer is evaluated by what remains useful after rebinding, not by whether the snapshot is always the best solver.

\subsection{Design ablations: are the learned layers decorative?}
To test whether the architecture earns its components rather than carrying them as decoration, we ablate each layer in isolation, holding everything else fixed. All are token-free re-runs of the real engine.

\begin{figure}[pos=!htbp]
\centering
\includegraphics[width=\columnwidth]{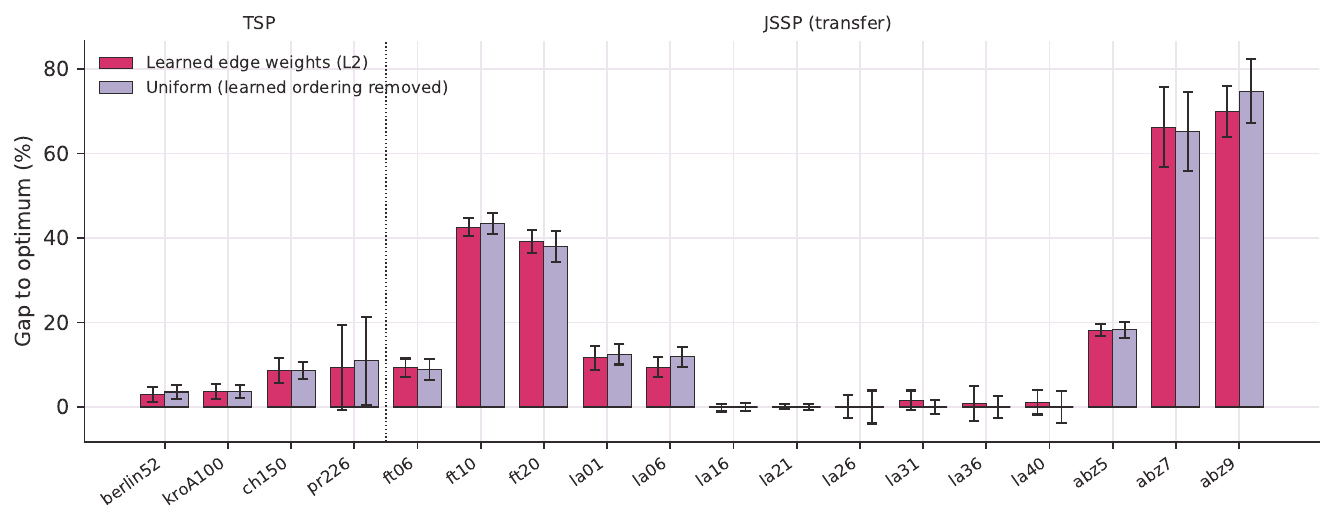}
\caption{L2 ablation. Learned operator edge weights versus uniform weights, holding operators, rules, and phases fixed. The learned ordering lowers the gap, modestly and most on the larger instances.}
\label{fig:l2_ablation}
\end{figure}

\textit{L2 (learned edge weights).} Replacing the learned operator edge weights with uniform weights---keeping the same operators, rules, and phases---removes the only quantity L2 learns. On TSP, the learned weighting lowers the gap on three of four instances, with the advantage largest on the biggest instance (pr226, $+1.58$ percentage points) and within noise on the mid-sized ones (Figure~\ref{fig:l2_ablation}). The effect is modest but consistent and grows with instance size, so the learned ordering helps within the source domain. On cross-domain JSSP transfer, however, the learned weighting is statistically indistinguishable from uniform (better on 8 of 14 instances, paired Wilcoxon $p=0.71$); the knowledge that transfers therefore resides in the operator pool and the ontology-constrained structure, not in the fine-grained learned weights.

\begin{figure}[pos=!htbp]
\centering
\includegraphics[width=0.72\columnwidth]{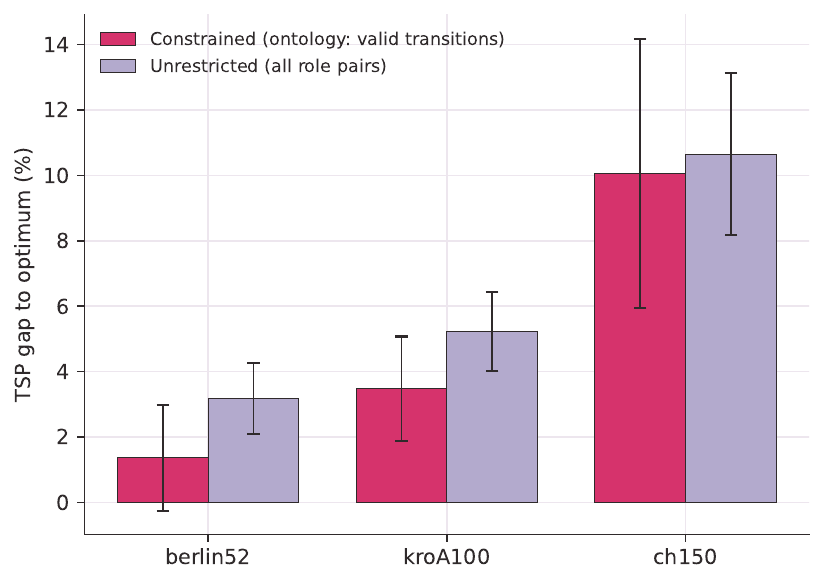}
\caption{L0 ontology ablation. The RoleSchema-constrained topology (valid category transitions only) versus an unrestricted topology (all role pairs), under the real ACO trained from scratch (eight seeds). The constrained topology lowers the mean TSP gap, so the ontology improves search rather than only restricting it.}
\label{fig:l0_ablation}
\end{figure}

\textit{L0 (ontology constraint).} We train the real ACO from scratch on the RoleSchema-constrained topology (only valid category transitions) versus an unrestricted topology (every role-to-role edge). The ontology constraint \emph{improves} search rather than merely limiting it: across eight seeds it lowers the mean TSP gap on all three converged instances (berlin52 $1.37$ versus $3.19$, kroA100 $3.47$ versus $5.23$ percentage points; Figure~\ref{fig:l0_ablation}), because the unrestricted graph spends ACO budget on semantically invalid transitions. The constrained snapshot also transfers competitively to job-shop scheduling, with its advantage concentrated on the largest instance. This answers whether the ontology helps or only restricts the search: it helps.

\begin{figure}[pos=!htbp]
\centering
\includegraphics[width=0.72\columnwidth]{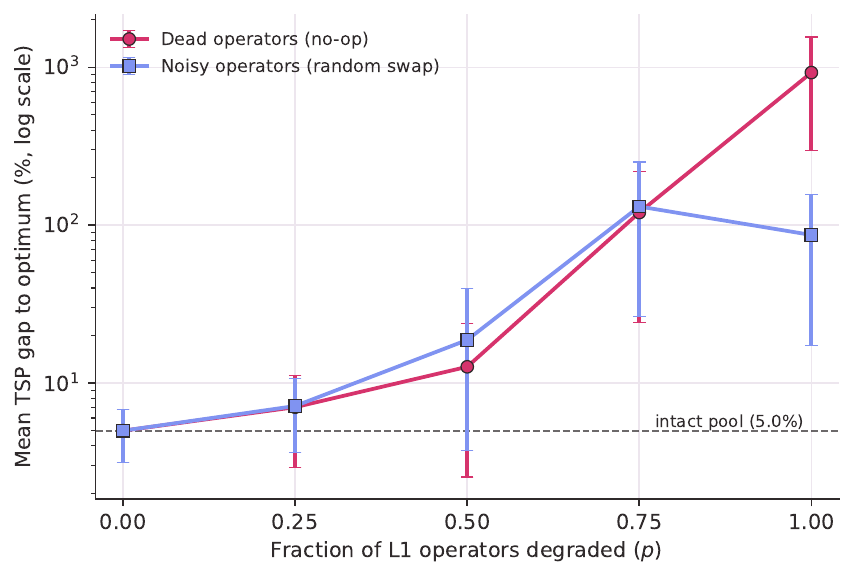}
\caption{L1 operator-quality ablation and robustness. Holding L0 and L2 fixed, a fraction $p$ of operator bodies is degraded to no-ops (\emph{dead}) or role-agnostic random swaps (\emph{noisy}); each degraded node retains its learned edge weight, so the learned policy still routes traffic to it. Mean TSP gap over four instances and eight seeds (log scale). A no-op pool collapses to the multistart baseline at $p=1$, confirming L1 is load-bearing, whereas a random-swap pool degrades far less because accept-if-improving traversal filters most harmful moves.}
\label{fig:l1_robustness}
\end{figure}

\textit{L1 (operator quality).} To isolate L1 and test robustness to low-quality operators, we hold L0 and L2 fixed and degrade a controlled fraction $p$ of the operator bodies; each degraded node keeps its identity and learned edge weight but loses its procedural content. Two regimes are tested over eight seeds and four TSP instances (Figure~\ref{fig:l1_robustness}): \emph{dead} operators become no-ops, and \emph{noisy} operators become role-agnostic random swaps. The intact pool reaches a $3.4$--$8.0\%$ gap. Replacing every operator with a no-op collapses the gap to $298$--$1978\%$: with nothing left to transform the solution, the executor is reduced to its multistart initial tours, which confirms that L1 content is load-bearing rather than decorative. Replacing every operator with a random swap is far less damaging ($33$--$206\%$), because the accept-if-improving traversal still extracts weak local search from random moves. Degradation is graceful at low corruption---at $p=0.25$ the gap stays within noise of the intact pool---but passes through a high-variance threshold near $p=0.75$, where the outcome depends on whether the few critical construction and local-search operators happen to survive. The architecture therefore tolerates a substantial fraction of low-quality operators before performance breaks down, while still depending on genuine L1 content for its results.

\subsection{GEAKG as a knowledge artifact}
The learned graph can be inspected independently of downstream task scores: the deployable knowledge core---typed topology, learned edge weights, and symbolic rules---is sparse and compact, storing tens of edges that serialize to kilobyte-scale JSON (about 1--3 KB for NAS and about 4 KB for the optimization core, separate from the larger offline construction history). A full structural, spectral, and rule-quality characterization is given in Methods (Artifact analysis). In brief, the normalized NAS transition operator has a spectral gap of 0.19 and stationary entropy 0.91, indicating a stable but non-collapsed traversal policy; redundancy elimination reduces 42 candidate transitions to 18 non-redundant rules carrying AMIE-style confidence and trace support; and dominant traversal paths expose reusable architecture-design sequences. The artifact perspective closes the loop with the transfer results: learned transition weights carry value within the source family, while cross-domain transfer is carried by the L0 and L1 ontology-constrained executable structure. Separating the stored artifact from any single task score is what keeps the representation inspectable and portable across operator sources, instance sizes, and baselines.

\section{Discussion}
\label{sec:discussion}
GEAKG reframes algorithm-design knowledge as an executable knowledge graph. The central contribution is representational: procedural know-how is stored as typed, executable graph structure rather than as isolated source files or transient search traces. This framing connects algorithm design to knowledge-graph concepts such as ontology constraints, learned edge confidence, rule extraction, and refinement~\cite{hogan2021knowledge,paulheim2017knowledge}. It also differs from code-evolution systems such as LLaMEA, FunSearch, EoH, and ReEvo, where knowledge is primarily implicit in generated programs~\cite{10752628,Romera-Paredes2024,10.5555/3692070.3693374,ye2024reevo}. In GEAKG, generated or hand-written operators become components inside a persistent graph.

The two case studies support this view at different transfer granularities. NAS demonstrates cross-dataset transfer within architecture families: the same procedural prior improves over random operator ordering across 70 transfer pairs and remains competitive with Regularized Evolution at zero deployment-token cost. Optimization demonstrates cross-domain transfer over a shared permutation representation. The ontology-constrained executable structure remains executable on JSSP and QAP, with the JSSP results showing that transfer is strongest on larger instances. These results support knowledge persistence and reuse while leaving room for stronger target-side solvers.

The RoleSchema is both a strength and a limitation. It gives the graph semantic structure, constrains invalid transitions, and allows the same engine to run in different domains. It also requires human modeling. In the present experiments, schema design took hours rather than days, and the cost is amortized at the level of a problem family: one 11-role optimization RoleSchema is reused unchanged across TSP, JSSP, and QAP (and binds to vehicle routing and bin packing). The L0 ablation shows that this schema contributes empirically by lowering the search gap rather than merely restricting it (Figure~\ref{fig:l0_ablation}). Future work should study LLM-assisted schema construction, schema ablations, and cases where multiple schemas compete for the same domain.

The evaluation has important limits. NAS uses tabular benchmarks; the tabulated 200-epoch values are rank-faithful across seeds, but full retraining under independent pipelines remains future work. Optimization transfer uses eight-seed JSSP runs and five-seed QAP runs over eight QAPLIB instances; broader QAP coverage remains future work. The current ablations isolate each layer individually---L0 ontology, L1 operator quality, and L2 sequencing---without a full factorial decomposition with interaction terms over the three layers. The graph also has no online adaptation. During deployment, the Symbolic Executor applies fixed transition weights and rules. Because it performs no weight update, sequential multi-domain transfer is non-destructive: the executor reads the snapshot read-only, and the learned edge weight state is byte-identical before and after a JSSP$\rightarrow$QAP$\rightarrow$JSSP deployment chain (verified by hashing). Catastrophic forgetting is therefore absent by construction, with online adaptation left to future work.

There are failure modes that require targeted study. Transfer can be negative when a target domain violates assumptions encoded in the source RoleSchema or learned edge weight distribution. Low-quality operators reduce performance even when the graph structure is valid, though the engine tolerates a substantial fraction of them before performance breaks down (Figure~\ref{fig:l1_robustness}). Operators bound to the same role are largely behaviorally consistent rather than semantically ambiguous: across construction and intensification roles they agree on the direction of their effect on most probe states (construction roles 100\%, small-intensification 93\% over 40 probes), and the only low-agreement role is perturbation, where behavioral diversity is the intended design. Conditional edges may become unreliable when runtime contexts differ from offline training traces. Incompatibility penalties rely on thresholds that are useful engineering choices, not yet theoretically optimized. These issues bound the claims and motivate robustness experiments against noisy operator pools, perturbed edge weights, and mismatched schemas.

The theoretical status is similarly conditioned. L2 learning occurs on a fixed graph and follows Max--Min Ant System dynamics~\cite{STUTZLE2000889}. This gives convergence behavior within the generated topology, without guaranteeing that the topology contains a globally optimal algorithmic path. The language model affects the offline search space by proposing topology and operators; validation and ACO then refine that frozen space. Conditional on a generated graph, L2 is standard MMAS and inherits its convergence behavior. The topology choice changes the fixed search space being converged in, and its magnitude is empirical rather than analytical. The L0 ablation bounds that effect: ontology-constrained topologies converge to lower gaps than unrestricted ones (Figure~\ref{fig:l0_ablation}). At deployment, no learning occurs, so convergence is an offline-construction property.

A practical advantage of GEAKG is lifecycle separation. Acquisition happens offline through generation and validation. Refinement happens through ACO over traces. Persistence happens through a compact snapshot. Application happens through symbolic execution without runtime model calls. This separation reduces deployment cost and provides an audit trail: roles, edges, operators, edge weights, and rules can be inspected separately.

Future work should expand the multi-seed transfer studies, add stronger domain-specific baselines such as adaptive large-neighborhood search and reinforcement-learning-guided heuristic selection, and evaluate real-training NAS deployments. It should also introduce formal query languages for procedural KGs, snapshot versioning for continual learning, and methods for detecting catastrophic transfer before deployment. The broader opportunity is to make algorithmic procedure a first-class knowledge artifact: generated, executed, evaluated, serialized, and reused as graph-structured knowledge.

\section{Methods}
\label{sec:methods}
\subsection{Formal GEAKG definition}
A GEAKG is a six-tuple $\mathcal{G}=(\mathcal{S},V,E,\Lambda,\Phi,\Sigma)$. The RoleSchema $\mathcal{S}=(\mathcal{R},\mathcal{K},\kappa,T,K_0,\mathrm{Re},M)$ is the ontology: $\mathcal{R}$ is the set of roles, $\mathcal{K}$ is the set of categories, $\kappa:\mathcal{R}\to\mathcal{K}$ assigns each role to a category, $T\subseteq\mathcal{K}\times\mathcal{K}$ defines admissible category transitions, $K_0$ defines entry categories, $\mathrm{Re}$ marks revisitable categories, and $M$ stores role metadata. The node set is $V=\mathcal{R}$. The edge set $E\subseteq V\times V$ is constrained by the ontology: $(v_i,v_j)\in E$ implies $(\kappa(v_i),\kappa(v_j))\in T$. The mapping $\Lambda:V\to2^{\mathcal{O}}$ assigns executable operators to each role. The function $\Phi:E\to\mathbb{R}^{+}$ stores learned edge weights. The rule set $\Sigma$ stores symbolic rules of the form condition-to-action over graph state, edge weights, and execution context.

A GEAKG is well formed when every edge satisfies RoleSchema constraints and every required category is reachable from an entry node. It is generative when topology and operators are synthesized by a language model, executable when every operator in $\Lambda(v)$ can be invoked by the runtime, and transferable when $\Phi$ and $\Sigma$ can be reused in a target domain with a compatible binding. \tcheck{Table~\ref{tab:roleschema_instances} lists the RoleSchema instantiations used in the two case studies.}

\begin{table}[pos=!htbp]
\centering
\caption{RoleSchema instantiations used in the two case studies.}
\label{tab:roleschema_instances}
\begin{tabular}{@{}lcc@{}}
\toprule
Component & NAS & Optimization \\
\midrule
Roles $|\mathcal{R}|$ & 18 & 11 \\
Categories $|\mathcal{K}|$ & 5 & 3 \\
Category transitions $|T|$ & 12 & 7 \\
Entry categories $|K_0|$ & 1 & 1 \\
Revisitable categories & 3/5 & 2/3 \\
Representative L0 edges & 42 & 14 \\
Directed density & 0.13 & 0.13 \\
\bottomrule
\end{tabular}
\end{table}

\subsection{Three-layer architecture and construction}
L0 is the MetaGraph: a typed graph over roles, admissible transitions, initial edge priors, and optional edge conditions. L1 is the executable operator pool. Each role has one or more operators that accept a domain context and return a candidate object or modified solution. Operators are validated before deployment through syntax checks, timeouts, and domain-specific validity tests. L2 is the learned procedural layer. ACO updates edge weights over role transitions using executed paths and observed fitness, while symbolic rules summarize high-confidence transitions and incompatibilities. In both case studies, the MetaGraph topology and the operator implementations are generated offline by a language model---gpt-5.2-0111 for the main runs, with gpt-4o-mini and the locally hosted Qwen2.5-14B as additional tiers in the optimization robustness study---at generation and repair temperatures 0.7 and 0.2--0.3 (Supplementary Table~S6). A generated operator is admitted to L1 only after execution-based validation on probe instances, and the frozen result of construction is serialized as the snapshot.

\begin{algorithm}[!htbp]
\caption{GEAKG construction}
\label{alg:geakg_construction}
\begin{algorithmic}[1]
\REQUIRE RoleSchema $\mathcal{S}$, training instances $\mathcal{I}$, language model $\mathcal{L}$, token budget $B$
\ENSURE Snapshot $\mathcal{G}=(L_0,L_1,L_2)$
\STATE Generate and validate L0 topology under category-transition constraints
\STATE Initialize L1 with base operators for every role
\WHILE{tokens used $<B$}
\STATE Generate candidate operators for selected roles and validate them
\STATE Run ACO over $L_0$ using operators from $L_1$ on $\mathcal{I}$
\STATE Update edge weights and symbolic incompatibility statistics
\ENDWHILE
\STATE Extract symbolic rules $\Sigma$ from high-confidence transitions
\RETURN $(L_0,L_1,(\Phi,\Sigma))$
\end{algorithmic}
\end{algorithm}

\subsection{ACO/MMAS formulation and convergence argument}
At each construction step, an ant at role $r_i$ selects a successor $r_j$ from the valid neighborhood $\mathcal{N}(r_i)$ with probability
\begin{equation}
P(r_j\mid r_i)=\frac{[\tau_{ij}]^{\alpha}[\eta_{ij}b_{ij}c_{ij}]^{\beta}}{\sum_{k\in\mathcal{N}(r_i)}[\tau_{ik}]^{\alpha}[\eta_{ik}b_{ik}c_{ik}]^{\beta}},
\end{equation}
where $\tau_{ij}$ is the learned edge weight, $\eta_{ij}$ is the L0 prior, $b_{ij}$ is a condition boost, and $c_{ij}$ is a compatibility factor. The condition boost multiplies the transition weight by a factor in $[1.5,2.5]$ when the runtime condition attached to the edge (an element of the rule set $\Sigma$, for example, a stagnation trigger) is satisfied, and equals $1$ otherwise. The compatibility factor equals $1$ by default and is lowered to $0.3$ for transitions with a persistent failure history, as specified next. We use Max--Min Ant System~\cite{STUTZLE2000889}: only the best ant augments edge weights, evaporation is applied each iteration, and edge weights are clipped to $[\tau_{\min},\tau_{\max}]$. The update is $\tau_{ij}\leftarrow(1-\rho)\tau_{ij}+\Delta\tau_{ij}^{\mathrm{best}}$, where $\Delta\tau_{ij}^{\mathrm{best}}=Q/f_{\mathrm{best}}$ when the best path uses edge $(i,j)$ and $0$ otherwise. Bounds with $\tau_{\min}>0$ keep all valid transitions selectable. L2 therefore inherits MMAS convergence behavior on the fixed L0 graph: edge weights are higher on high-quality paths permitted by the graph while no edge is permanently removed.

Each ant has energy $E\sim\mathcal{U}(4,12)$ controlling path length. Incompatibility statistics reduce the compatibility factor when a transition appears in more than 30\% of failures after at least 10 samples. The Symbolic Executor deploys the snapshot by repeatedly selecting a role using $\Phi$, $\Sigma$, and execution context, binding it to an operator in $\Lambda(r)$, applying it through the domain context, and validating the result. No language-model calls are made during deployment.

\subsection{RoleSchema design methodology}
We design a RoleSchema through five steps: domain analysis, operator-concept enumeration, category grouping, property definition, and transition-constraint specification. In optimization, the ILS literature motivates a construction--intensification--perturbation skeleton~\cite{lourenco2003iterated}. This yields 11 roles grouped into Construction, Local Search, and Perturbation. Construction is the entry category; Local Search and Perturbation are revisitable; category-level transitions define which operator classes may follow each other. In NAS, the literature on DARTS, ENAS, NASNet, Once-for-All, and Regularized Evolution motivates roles for topology, activation, training, regularization, and evaluation~\cite{liu2019darts,pham2018efficient,zoph2018learning,cai2020once,real2019regularized}. Category-level constraints keep schema construction scalable: the transition table has $O(|\mathcal{K}|^2)$ entries, and each new role inherits the transition rules of its category.

\subsection{Experimental protocol and complexity}
Optimization experiments use TSP as the source domain and JSSP/QAP as transfer targets. JSSP instances include Fisher--Thompson, Lawrence, Adams--Balas--Zawack, and Taillard benchmarks~\cite{fisher1963probabilistic,lawrence1984resource,adams1988shifting,taillard1993benchmarks}. The JSSP transfer evaluation uses eight seeds with a 30-second limit per instance and precedence-aware makespan evaluation. The from-scratch ILS baselines are our own implementations of the standard iterated-local-search scheme~\cite{lourenco2003iterated}: first-improvement local search over adjacent swaps of the permutation, restarted through random perturbations of the incumbent, instantiated per target domain (JSSP, QAP, and linear ordering) and provided in the accompanying code. NAS experiments use NAS-Bench-Graph and NAS-Bench-201. Each NAS setting uses 10 runs with seeds 42--51 and 200 architecture evaluations per run. Baselines are Random Search and Regularized Evolution; Bayesian Optimization is reported in the NAS deployment-cost Methods subsection as an additional scalability reference. Statistical tests use Wilcoxon signed-rank at $\alpha=0.05$.

Let $R=|\mathcal{R}|$, $|E|$ be learned edges, $k$ generated operators per role, $A$ ants per iteration, $T$ ACO iterations, $\bar{L}$ mean traversal length, $\bar{d}$ mean out-degree, and $c_v$ validation cost. L0 topology generation considers $O(R^2)$ candidate role pairs but uses one structured generation. L1 synthesis and validation cost $O(Rkc_v)$ plus $O(Rk)$ language-model calls. L2 ACO costs $O(|E|AT)$ and uses no language-model calls. Online symbolic execution costs $O(\bar{L}\bar{d})$ per generated solution and uses no language-model calls. Snapshot space is $O(|E|+R)$ for the deployable knowledge core.

\subsection{Artifact analysis: structure, spectra, and rules}
This section reports in full the artifact characterization summarized in Section~\ref{sec:results} (Results), covering structural statistics, edge weight spectra, and symbolic-rule quality for the learned NAS and optimization snapshots.

The learned graph can be inspected independently of downstream task scores. The NAS and optimization graphs are sparse, typed, and compact: representative snapshots store tens of edges and serialize to kilobyte-scale JSON. The relevant object is the deployable knowledge core, consisting of topology, edge weights, and symbolic rules; operator libraries and offline provenance can be stored separately.

The structural statistics show that the learned artifacts are small enough to inspect directly. The NAS case uses 18 roles in 5 categories; the optimization case uses 11 roles in 3 categories. The learned NAS graphs contain 32--50 edges, with directed density 0.105--0.163 and average out-degree 1.8--2.8. The optimization graph contains 14 generated edges (whose weights are then learned) over 11 roles, with directed density 0.13 and average out-degree 1.3. The corresponding symbolic rule sets contain 10 and 8 rules, and the deployable snapshots are about 1--3 KB for NAS and about 4 KB for the optimization knowledge core (the full optimization snapshot, including offline construction history, is about 17 KB, generated at a one-time cost of about 50k language-model tokens). These sizes matter because they distinguish the deployed knowledge core from the larger offline construction process. The graph is not a long trace log; it is a compact set of typed transitions, weights, rules, and operator bindings.

Sparsity is also semantically meaningful. A fully connected directed graph over roles would allow many role sequences that violate the domain ontology. GEAKG instead restricts edges through the RoleSchema and then lets ACO refine the admissible transitions. In NAS, the baseline RoleSchema topology has 32 edges over 18 roles, while LLM-generated topologies reach 42 edges for gpt-5.2 and 50 edges for gpt-4o-mini. Variation across datasets then appears in the learned weights rather than in a different role vocabulary. This structure keeps the graph interpretable: a high-weight edge can be read as evidence for a specific procedural transition under an explicit category constraint.

The two graphs are comparably sparse: the optimization graph has 14 generated edges (with learned weights) over 11 roles, close in directed density to the NAS graph, because only semantically valid category transitions are realized. Their shapes still differ in a useful way. In local-improvement search, construction, intensification, and perturbation can often be revisited in multiple orders, while NAS has a more pipeline-like ordering from topology to activation to training and regularization. The difference is useful because it shows that GEAKG does not impose one graph shape. The RoleSchema defines admissible categories; L0 realizes a graph within those constraints; L2 then concentrates probability mass on the transitions that perform well during execution.

\begin{figure}[pos=!htbp]
\centering
\includegraphics[width=0.8\columnwidth]{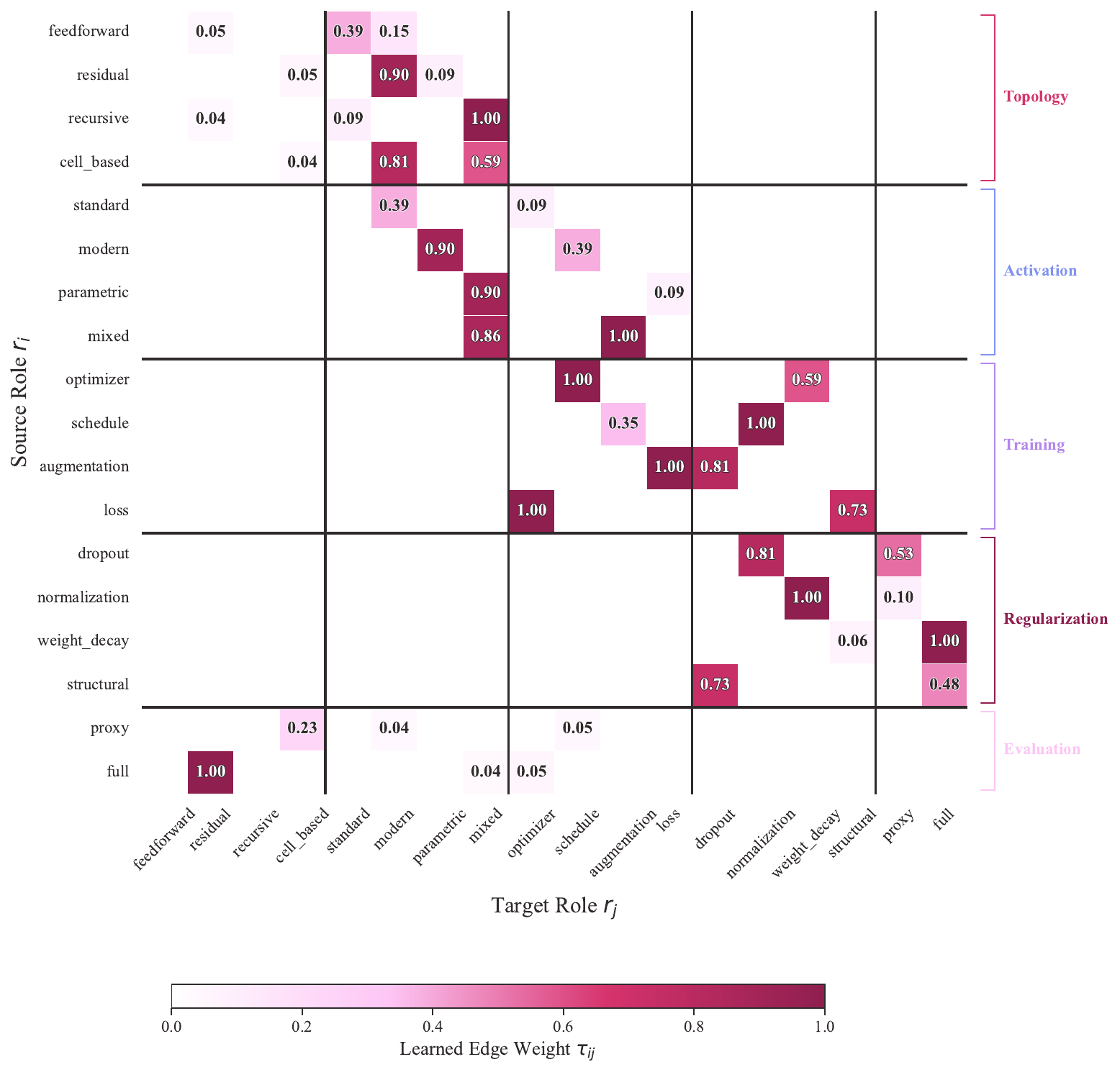}
\caption{Learned edge-weight matrix for the NAS GEAKG on Cora. Darker entries correspond to reinforced role transitions. The block structure follows RoleSchema categories and exposes the learned architecture-design pipeline.}
\label{fig:kg_pheromone_heatmap}
\end{figure}

Edge weights reveal learned procedural structure. Figure~\ref{fig:kg_pheromone_heatmap} shows the NAS Cora matrix, where high-weight transitions form a pipeline from topology through activation, training, regularization, and evaluation, with feedback edges that support iterative refinement. Spectral analysis of the normalized transition operator gives a second-largest eigenvalue modulus of 0.81, hence a spectral gap of 0.19. The stationary distribution remains spread across the pipeline, with normalized entropy 0.91 and mean per-row transition entropy 0.71. These quantities indicate a stable but non-collapsed transition policy.

The per-edge distribution shows that ACO pushes the graph toward a structured policy while MMAS bounds prevent collapse. In the NAS Cora GEAKG, 9 of the 42 edges, or 21\%, are saturated at $\tau_{\max}=1.0$, while 14 edges, or 33\%, approach $\tau_{\min}$. The full edge weight range spans 0.04 to 1.0, with a mean of 0.51. At the aggregate level, learned distributions achieve 3--4\% entropy reduction from the uniform maximum (Supplementary Fig.~S5). This reduction is conservative because MMAS intentionally bounds edge weights; under those bounds, the theoretical maximum entropy reduction is approximately 15--20\%, so the observed 3--4\% represents a nontrivial fraction of the available concentration.

The spectral attractor gives a complementary view of the same learned structure. After normalizing outgoing edge weights into a row-stochastic transition matrix, the dominant stationary role is \texttt{act\_mixed}, with stationary mass $\pi=0.14$. The attractor is therefore centered on an activation-composition role but remains distributed across the design pipeline. This matches the rule-quality table, where \texttt{act\_mixed} appears as a frequent consequent and antecedent. The spectral gap of 0.19 indicates that the transition dynamics mix rather than oscillate, while the high stationary entropy indicates that traversal retains multiple viable paths. A direct perturbation test corroborates this stability: perturbing the learned edge weights by a magnitude $\varepsilon$ shifts the executor output by only a small amount that grows smoothly and remains bounded as $\varepsilon$ increases, with no discontinuous jumps (Supplementary Figure~S1); the learned policy therefore does not bifurcate under perturbation.

Symbolic rules provide another view of the graph. Table~\ref{tab:rule_quality} reports representative non-redundant NAS rules with AMIE-style confidence and trace support~\cite{galarraga2013amie}. Confidence is the normalized outgoing edge weights $P(r_j\mid r_i)$; support is the number of occurrences in top execution traces. High-confidence low-support rules and lower-confidence high-support rules encode different forms of procedural evidence, which is why both metrics are reported.

\begin{table}[pos=!htbp]
\centering
\caption{Representative non-redundant rules from the NAS Cora GEAKG with confidence and trace support.}
\label{tab:rule_quality}
\small
\begin{tabular}{@{}llcc@{}}
\toprule
Antecedent ($r_i$) & Consequent ($r_j$) & Conf. & Support \\
\midrule
\texttt{reg\_weight\_decay} & \texttt{eval\_full} & 0.94 & 0 \\
\texttt{act\_parametric} & \texttt{act\_mixed} & 0.91 & 9 \\
\texttt{topo\_recursive} & \texttt{act\_mixed} & 0.88 & 3 \\
\texttt{topo\_residual} & \texttt{act\_modern} & 0.86 & 4 \\
\texttt{act\_modern} & \texttt{act\_parametric} & 0.70 & 7 \\
\texttt{reg\_structural} & \texttt{reg\_dropout} & 0.60 & 6 \\
\texttt{train\_augmentation} & \texttt{train\_loss} & 0.55 & 10 \\
\texttt{act\_mixed} & \texttt{train\_augmentation} & 0.54 & 15 \\
\bottomrule
\end{tabular}
\end{table}

The rule set also shows that the graph can be summarized without replaying every trace. Redundancy elimination reduces the 42 candidate transitions in the NAS Cora graph to 18 non-redundant rules by keeping the dominant consequent for each antecedent when confidence is at least 0.5. Four of the 18 non-redundant rules reach confidence of at least 0.9. Others have lower confidence but broader support, such as \texttt{act\_mixed} $\to$ \texttt{train\_augmentation}, with confidence 0.54 and support 15. This distinction is useful for auditability: a rule with high confidence but low support may be a sharp local preference, while a moderate-confidence high-support rule records a common transition that remains useful across multiple paths.

Together, the artifact analyses show that GEAKG stores typed roles, admissible transitions, executable operator bindings, learned edge strengths, and rule summaries. This makes the procedural representation inspectable and portable, even when the operators originate from language-model generation or code-evolution methods.

Dominant traversal paths make the stored procedure even more concrete (Supplementary Fig.~S6). In the NAS Cora graph, the top paths follow the category ordering Topology $\to$ Activation $\to$ Training $\to$ Regularization, sometimes extending to Evaluation. Reported path lengths range from 6 to 9 roles, with 6-role paths most frequent. The most common path is \texttt{topo\_residual} $\to$ \texttt{act\_modern} $\to$ \texttt{act\_parametric} $\to$ \texttt{act\_mixed} $\to$ \texttt{train\_augmentation} $\to$ \texttt{train\_loss}, with frequency $n=4$. This path records a reusable architecture-design sequence: start from residual topology, refine activation choices, and then apply augmentation-driven training. The important point is not that this single path is universally optimal. It is that GEAKG exposes such paths as explicit procedural objects that can be audited, compared, and transferred.

The artifact perspective closes the loop with the empirical transfer results. NAS transfer is supported by correlated edge weights across datasets, lower variance than a per-target evolutionary baseline, and dominant paths that follow a coherent architecture-design order. Optimization transfer is supported by a compact graph whose roles are generic enough to act on routing, scheduling, and assignment permutations. In both cases, the graph is small, structured, and executable. The learned weights are part of the stored procedural artifact the Symbolic Executor applies at deployment; the ablations show they carry value within the source family, while cross-domain transfer is carried by the L0 and L1 structure.

This distinction between artifact and outcome is important for interpreting the experiments. A single task score can rise or fall with operator quality, instance size, or baseline choice. The GEAKG artifact remains analyzable across those conditions: its edges state which role transitions are admissible, its edge weights quantify learned preference, its rules summarize repeated transitions, and its paths expose executable procedures. That is why we report downstream performance together with this structural and spectral analysis. The empirical story is not only that the graph works on selected targets, but that the stored object has the properties expected of a procedural knowledge graph: compactness, executability, transferability, and inspectability.

\subsection{NAS transfer protocol, deployment cost, and stability}
This section reports the NAS transfer-design choices and the deployment-cost and stability measurements summarized in Section~\ref{sec:results} (Results).

The transfer layout is deliberately asymmetric across the two benchmarks because the benchmark datasets differ. NAS-Bench-Graph contributes 64 configurations: 8 source datasets, excluding Proteins as a source, crossed with target datasets and including self-transfer cases. NAS-Bench-201 contributes 6 configurations from 3 sources and 2 targets. This means the combined 70-pair Random Search result is dominated by the graph benchmark, while the CNN benchmark serves as a smaller cross-family check. For this reason, RegEvo comparisons are stated by benchmark rather than collapsed into an aggregate win count. The graph benchmark provides enough pairs to estimate a median delta against RegEvo; the cell benchmark mainly shows that the same procedural roles remain usable when the architecture encoding changes.

The learned NAS edge weights are also stable across datasets. In the NAS-Bench-Graph study, pairwise Pearson correlations of learned edge weights across all 9 datasets yield a mean correlation of $\bar{r}=0.91$ over 36 pairs, with all correlations significant at $p<0.001$. The range is 0.82 for Photo versus Proteins to 0.95 for Computers versus Cora. This matters because cross-dataset transfer depends on the same role transitions being reinforced under different data distributions. The high correlation indicates that the learned weights are not specific to one source dataset; they encode a consistent ordering over architecture-design roles that remains correlated across graph-learning tasks.

The Symbolic Executor also has a variance advantage over Regularized Evolution on representative transfer pairs. Across the reported pairs, its standard deviation is 1.2$\times$--4.8$\times$ lower than Regularized Evolution (Supplementary Fig.~S4). On ImageNet16-120$\to$CIFAR-100, the standard deviation is 0.10 for the Symbolic Executor and 0.49 for Regularized Evolution, a 4.8$\times$ reduction, while the mean accuracies remain close at 73.47 and 73.34, respectively. This is not a peak-performance claim. It indicates that a fixed procedural snapshot can provide repeatable behavior, which is useful when the cost of repeated architecture-search runs is nontrivial.

The scalability comparison with Bayesian Optimization reinforces the deployment-cost point. Using Gaussian Process Expected Improvement from scikit-optimize~\cite{scikit-optimize}, the original experiments compare short-budget and unlimited BO settings. GEAKG completes 500 evaluations in under 7 seconds. On NAS-Bench-201, GEAKG reaches 71.42\% mean accuracy in approximately 4.1 seconds for 500 evaluations; short-budget BO reaches 69.78\% in 2.1 seconds with approximately 28 evaluations; unlimited BO reaches 71.59\% in approximately 1350 seconds for 500 evaluations. On NAS-Bench-Graph, GEAKG reaches 75.70\% in approximately 0.3 seconds; short-budget BO reaches 25.27\% in 0.5 seconds with approximately 22 evaluations; unlimited BO reaches 76.20\% in approximately 1302 seconds. BO can match or slightly exceed GEAKG when allowed the full expensive run, but it spends computation per target query. GEAKG pays its construction cost offline and then deploys the stored graph.

This cost structure is central to the interpretation of the NAS results. The 70 transfer pairs, with 10 runs each, complete in under 140 seconds once the snapshot exists. Per transfer, the Symbolic Executor takes approximately 0.1 seconds on NAS-Bench-Graph and 1.8 seconds on NAS-Bench-201, with zero deployment tokens. The snapshot is approximately 2 KB in the NAS case. These numbers are small compared with the offline L1 construction budget and are the reason parity with a stronger per-target search method is meaningful. The result is not that graph traversal is always the most accurate NAS optimizer; it is that reusable procedural knowledge can remove most of the marginal search cost while preserving competitive accuracy.

\section*{Data availability}
The NAS benchmarks (NAS-Bench-Graph; NAS-Bench-201/NATS-Bench), the TSP instances (TSPLIB), and the JSSP instances (Fisher--Thompson, Lawrence, Adams--Balas--Zawack) are public. The QAP instances are from QAPLIB; the linear-ordering instances are reduced-size variants of LOLIB matrices, both provided in the repository. The per-experiment result files (E1--E6) behind every reported number are included in the archived artifact.

\section*{Code availability}
All code, the canonical learned snapshot, the operator pools, RoleSchemas, and prompt templates are archived in a versioned snapshot on Zenodo (\href{https://doi.org/10.5281/zenodo.21940090}{doi:10.5281/zenodo.21940090}). Each headline number can be regenerated with the documented \texttt{uv run python scripts/<\dots>} command for its experiment; sampling temperatures, model versions, seeds, and validation heuristics are listed in Supplementary Table~S6.

\section*{Acknowledgements}
C.C.S. acknowledges funding from R\&D Project PID2022-138283NB-I00, funded by MICIU/AEI/10.13039/501100011033 and by ``FEDER -- A way of making Europe''. C.B. was supported by grant PID2022-136787NB-I00, funded by MCIN/AEI/10.13039/501100011033, and acknowledges the support of his employer, the Spanish National Research Council (CSIC).

\section*{Author contributions}
C.C.S.: conceptualization, methodology, software, writing---original draft. J.H.G.: methodology, supervision. A.V.T.: methodology, supervision. C.B.: supervision, writing---review and editing.

\section*{Competing interests}
The authors declare no competing interests.


\clearpage
\section*{Supplementary Information}
\label{sec:supplementary_information}

\begin{center}
\large\textbf{Supplementary Information for:}\\[2pt]
\large Transferable knowledge graphs with executable learned operators for algorithm design\\[4pt]
Camilo Chac\'on Sartori, Jos\'e H. Garc\'ia, Andrei Voicu Tomut, Christian Blum
\end{center}

This file contains Supplementary Tables S1--S6 and Supplementary Figures S1--S6. They are organized under two Supplementary Notes. The items are numbered independently of the main text, and each item is cited there.

\setcounter{table}{0}
\setcounter{figure}{0}
\renewcommand{\thetable}{S\arabic{table}}
\renewcommand{\thefigure}{S\arabic{figure}}
\renewcommand{\tablename}{Supplementary Table}
\renewcommand{\figurename}{Supplementary Figure}

\subsection*{Supplementary Note 1: Source-domain, transfer, and reproducibility tables}
Supplementary Tables~\ref{tab:robustness}--\ref{tab:hyperparams} support the optimization case study and its reproducibility. Supplementary Tables~\ref{tab:robustness} and~\ref{tab:geakg_vs_llamea} report the TSP source-domain results that motivate transfer. These tables cover robustness across language-model tiers and a head-to-head comparison with LLaMEA under a matched 50k-token budget. Supplementary Table~\ref{tab:transfer_jssp} gives the per-instance TSP$\to$JSSP transfer makespans. The table uses the precedence-aware evaluation described in the main text. Supplementary Tables~\ref{tab:qap_regenerated_partial} and~\ref{tab:qap-transfer} document the quadratic-assignment boundary. They include the learned-versus-uniform-versus-ILS ablation and an earlier reference evaluation over a wider set of QAP instances, retained for completeness. Supplementary Table~\ref{tab:hyperparams} lists the ACO/MMAS, incompatibility, language-model, and protocol hyperparameters needed to reproduce every run.

\subsection*{Supplementary Tables}
\begin{table}[pos=!htbp]
\centering
\caption{TSP performance with mid-tier model gpt-4o-mini. Gap (\%) is mean$\pm$std over 15 runs. A dash means the method returned no valid tour within the token budget.}
\label{tab:robustness}
\small
\begin{tabular}{lrrrrr}
\toprule
Instance & $n$ & GEAKG (10k) & GEAKG (50k) & LLaMEA (50k) & Winner \\
\midrule
berlin52 & 52 & $1.99\pm1.03$ & $2.13\pm0.99$ & $0.03\pm0.00$ & LLaMEA \\
kroA100 & 100 & $2.24\pm0.99$ & $2.05\pm0.91$ & $0.55\pm0.14$ & LLaMEA \\
ch150 & 150 & $6.21\pm1.25$ & $6.11\pm1.45$ & $2.37\pm0.02$ & LLaMEA \\
pr226 & 226 & $2.17\pm0.55$ & $2.39\pm0.66$ & $1.46\pm0.82$ & LLaMEA \\
pcb442 & 442 & $13.25\pm1.03$ & $14.41\pm1.77$ & --- & GEAKG \\
rat783 & 783 & $50.10\pm2.03$ & $49.22\pm1.75$ & --- & GEAKG \\
pr1002 & 1002 & $54.95\pm3.06$ & $55.02\pm2.92$ & --- & GEAKG \\
\bottomrule
\end{tabular}
\end{table}

\begin{table}[pos=!htbp]
\centering
\caption{GEAKG versus LLaMEA on TSP with gpt-5.2 and 50k total token budget each. Gap (\%) is mean$\pm$std over 15 runs.}
\label{tab:geakg_vs_llamea}
\small
\begin{tabular}{@{}lrccc@{}}
\toprule
Instance & $n$ & GEAKG & LLaMEA & Improvement \\
\midrule
berlin52 & 52 & $0.031\pm0.00$ & $0.031\pm0.00$ & --- \\
kroA100 & 100 & $0.025\pm0.02$ & $0.016\pm0.00$ & --- \\
ch150 & 150 & $0.578\pm0.18$ & $0.828\pm0.26$ & 30\% \\
pr226 & 226 & $0.628\pm0.27$ & $1.810\pm0.00$ & 65\% \\
pcb442 & 442 & $3.444\pm0.37$ & $7.408\pm0.00$ & 54\% \\
rat783 & 783 & $9.158\pm2.89$ & $12.246\pm0.16$ & 25\% \\
pr1002 & 1002 & $8.880\pm2.41$ & $12.197\pm0.65$ & 27\% \\
\bottomrule
\end{tabular}
\end{table}

\begin{table}[pos=!htbp]
\centering
\caption{TSP $\to$ JSSP transfer (eight-seed means) under precedence-aware evaluation. Time limit: 30 seconds per instance. SPT and LPT denote the shortest- and longest-processing-time dispatching rules. Makespans are reported for all instances; gaps to the best-known makespan are given in parentheses where available.}
\label{tab:transfer_jssp}
\small
\begin{tabular}{lrrrrr}
\toprule
Instance & Ops & GEAKG & ILS & Best simple baseline & Winner \\
\midrule
ft06 & 36 & 60 (9.1\%) & 57 (3.6\%) & SPT 109 (98.2\%) & ILS \\
ft10 & 100 & 1348 (44.9\%) & 1350 (45.2\%) & SPT 2648 (184.7\%) & GEAKG \\
ft20 & 100 & 1633 (40.2\%) & 1416 (21.5\%) & LPT 2580 & ILS \\
la01 & 50 & 747 (12.2\%) & 682 (2.4\%) & SPT 1462 (119.5\%) & ILS \\
la06 & 75 & 1023 (10.5\%) & 1007 (8.7\%) & SPT 2367 (155.6\%) & ILS \\
la16 & 100 & 1199 & 1286 & LPT 3115 & GEAKG \\
la21 & 150 & 1486 & 1917 & SPT 4361 & GEAKG \\
la26 & 200 & 1881 & 2801 & LPT 5664 & GEAKG \\
la31 & 300 & 2479 & 4312 & SPT 8061 & GEAKG \\
la36 & 225 & 1976 & 2968 & LPT 6710 & GEAKG \\
la40 & 225 & 2005 & 3043 & SPT 5952 & GEAKG \\
abz5 & 100 & 1485 (20.3\%) & 1675 (35.7\%) & LPT 4658 & GEAKG \\
abz7 & 300 & 1149 (75.2\%) & 2210 (236.9\%) & SPT 4010 (511.3\%) & GEAKG \\
abz9 & 300 & 1164 (71.7\%) & 2160 (218.6\%) & LPT 4206 & GEAKG \\
\bottomrule
\end{tabular}
\end{table}

\begin{table}[pos=!htbp]
\centering
\caption{QAP transfer ablation. The table compares learned operator edge weights, uniform operator edge weights, and a from-scratch ILS over eight QAPLIB instances at five seeds each (mean gap to the known optimum, \%). Learned is indistinguishable from uniform (lower mean on 4/8 instances, paired Wilcoxon $p=0.64$). Thus, the L2 weights do not transfer to QAP. A from-scratch ILS is stronger than the transferred snapshot on every instance.}
\label{tab:qap_regenerated_partial}
\small
\begin{tabular}{lrrrr}
\toprule
Instance & $n$ & Learned & Uniform & ILS \\
\midrule
nug12 & 12 & 3.32 & 3.04 & 0.00 \\
nug20 & 20 & 2.61 & 4.14 & 0.14 \\
nug30 & 30 & 2.33 & 2.89 & 0.47 \\
chr15a & 15 & 17.77 & 17.73 & 0.32 \\
chr20a & 20 & 24.62 & 25.24 & 7.41 \\
chr25a & 25 & 47.54 & 45.69 & 6.24 \\
tai20a & 20 & 3.73 & 4.89 & 1.27 \\
tai50a & 50 & 4.27 & 3.95 & 3.05 \\
\bottomrule
\end{tabular}
\end{table}

\begin{table}[pos=!htbp]
\centering
\caption{QAP transfer reference results over a wider set of QAPLIB instances, obtained under an earlier evaluation protocol (mean$\pm$std over repeated runs) and retained for completeness. GL is the deterministic Gilmore--Lawler lower bound. The definitive learned-versus-uniform-versus-ILS comparison is Supplementary Table~\ref{tab:qap_regenerated_partial}.}
\label{tab:qap-transfer}
\small
\begin{tabular}{lrrrrr}
\toprule
Instance & $n$ & GEAKG & ILS & GL & Winner \\
\midrule
nug12 & 12 & $0.00\pm0.00$ & $0.00\pm0.00$ & 25.26 & Tie \\
nug15 & 15 & $0.00\pm0.00$ & $0.00\pm0.00$ & 28.00 & Tie \\
nug20 & 20 & $0.09\pm0.18$ & $0.00\pm0.00$ & 33.46 & ILS \\
nug25 & 25 & $0.25\pm0.20$ & $0.06\pm0.07$ & 36.06 & ILS \\
nug30 & 30 & $0.93\pm0.46$ & $0.56\pm0.15$ & 29.59 & ILS \\
tai20a & 20 & $1.47\pm0.46$ & $0.62\pm0.27$ & 30.15 & ILS \\
tai50a & 50 & $3.91\pm0.40$ & $3.29\pm0.47$ & 18.93 & ILS \\
tai80a & 80 & $4.96\pm1.04$ & $3.74\pm0.32$ & 16.27 & ILS \\
tai100a & 100 & $6.47\pm2.00$ & $4.09\pm0.25$ & 14.20 & ILS \\
tai150b & 150 & $7.05\pm0.63$ & $13.79\pm0.88$ & 30.36 & GEAKG \\
tai256c & 256 & $3.73\pm3.73$ & $17.18\pm2.20$ & 120.48 & GEAKG \\
\bottomrule
\end{tabular}
\end{table}

\begin{table}[pos=!htbp]
\centering
\caption{Hyperparameters used across experiments.}
\label{tab:hyperparams}
\small
\begin{tabular}{@{}lll@{}}
\toprule
Group & Parameter & Value \\
\midrule
ACO/MMAS & pheromone exponent $\alpha$ & 2.0 \\
 & heuristic exponent $\beta$ & 2.0 \\
 & evaporation $\rho$ & 0.1 \\
 & deposit constant $Q$ & 100 \\
 & ants per iteration & 10 \\
 & pheromone bounds $[\tau_{\min},\tau_{\max}]$ & $[0.001,1.0]$ \\
 & ant energy $E$ & $\mathcal{U}(4,12)$ \\
 & stagnation reset & 20 iterations \\
 & deposit rule & MMAS best ant only \\
Incompatibility & failure threshold & 0.30 \\
 & minimum samples & 10 \\
 & penalty factor & 0.30 \\
LLM generation & models & gpt-5.2-0111, gpt-4o-mini, Qwen2.5-14B \\
 & temperature generation/repair & 0.7 / 0.2--0.3 \\
Protocol & runs optimization/NAS & 15 / 10 (seeds 42--51 for NAS) \\
 & evaluations per NAS run & 200 \\
 & token budget & 10k--50k offline; 0 at deployment \\
\bottomrule
\end{tabular}
\end{table}

\subsection*{Supplementary Note 2: Artifact stability and structure figures}
\begin{figure}[pos=!htbp]
\centering
\includegraphics[width=0.72\columnwidth]{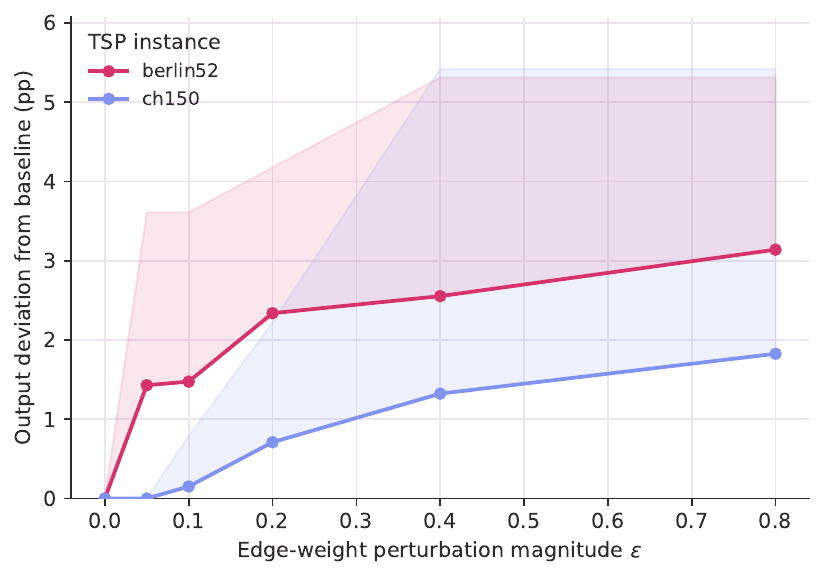}
\caption{Perturbation sensitivity (bifurcation test). The plot shows output deviation of the Symbolic Executor as the learned edge weights are perturbed by magnitude $\varepsilon$ (mean over random perturbations; shaded band to the maximum). The response is smooth and bounded. It has no discontinuous jumps, which indicates a stable learned policy.}
\label{fig:bifurcation}
\end{figure}

\begin{figure}[pos=!htbp]
\centering
\includegraphics[width=\textwidth]{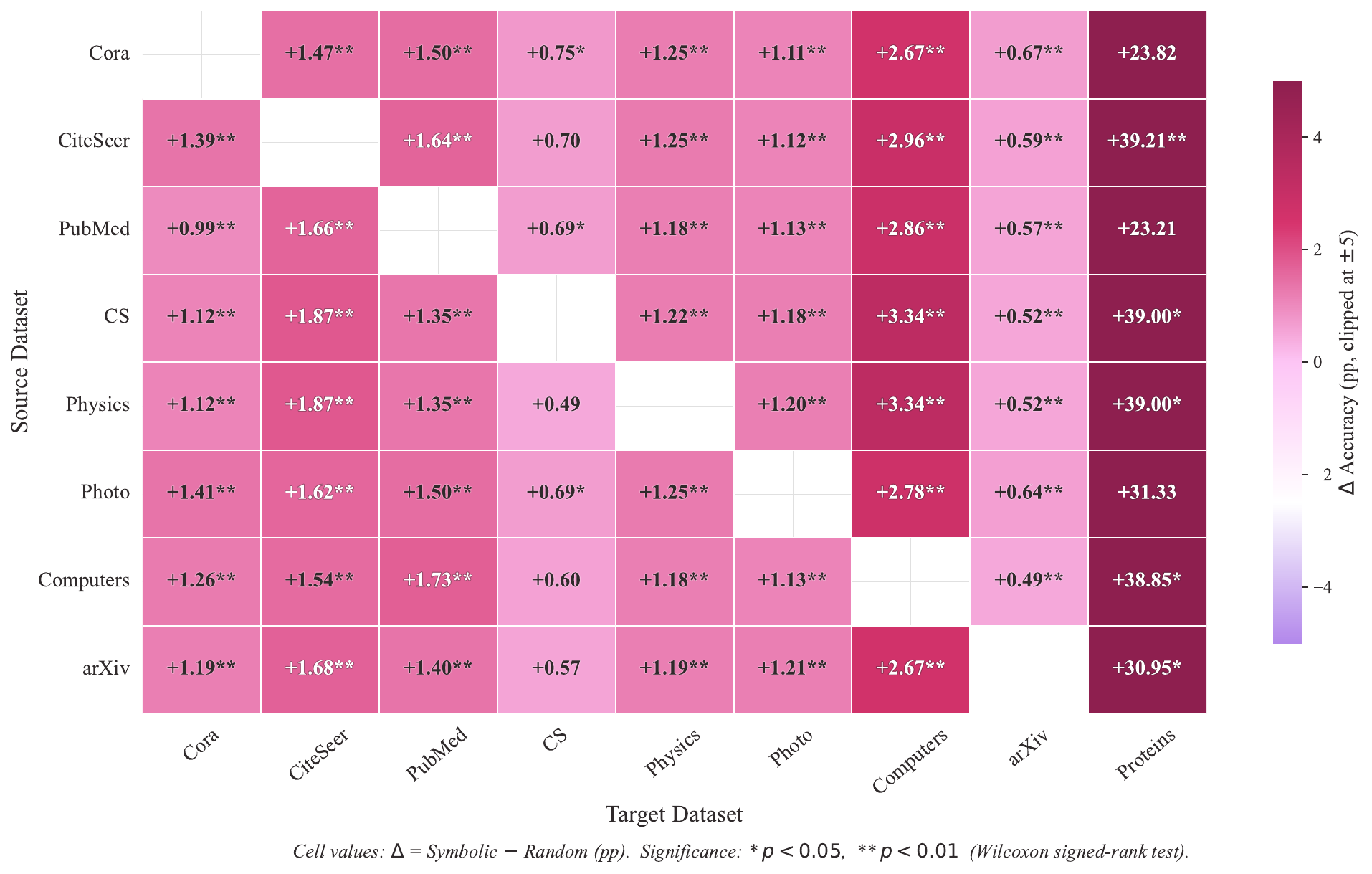}
\caption{Cross-dataset transfer on NAS-Bench-Graph. Each cell reports the test-accuracy difference (percentage points) between the GEAKG Symbolic Executor and Random Search for a source$\to$target dataset pair under a matched evaluation budget. Rows are source datasets, and columns are target datasets. Warmer cells favour GEAKG. All 64 pairs are positive, matching the 64/64 improvement reported in the main text.}
\label{fig:nas_heatmap_graph}
\end{figure}

\begin{figure}[pos=!htbp]
\centering
\includegraphics[width=\textwidth]{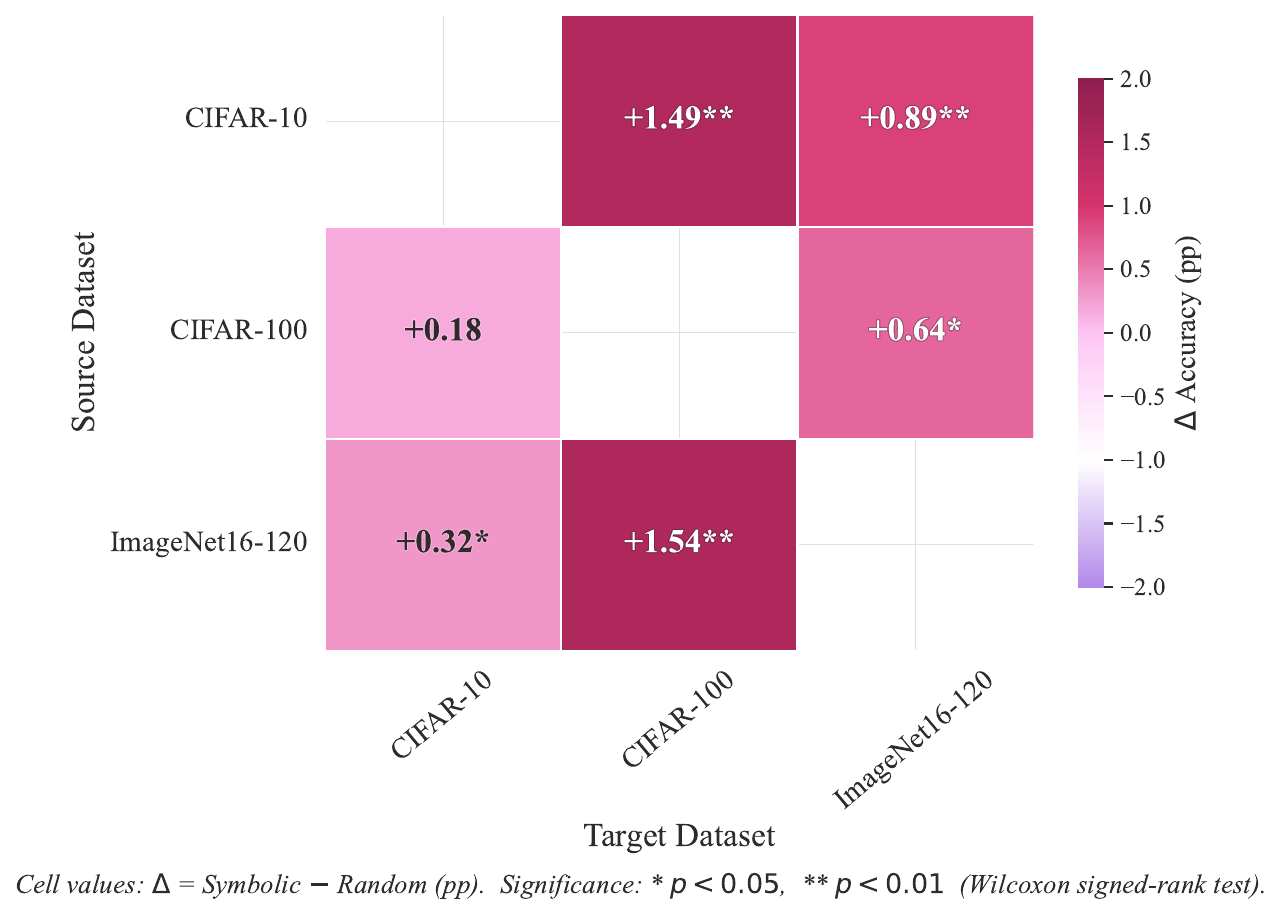}
\caption{Cross-dataset transfer on NAS-Bench-201. This panel follows Supplementary Fig.~\ref{fig:nas_heatmap_graph} for the six NAS-Bench-201 transfer pairs over CIFAR-10, CIFAR-100, and ImageNet16-120. Each cell reports the GEAKG-minus-Random-Search test-accuracy difference (percentage points). Positive values indicate cases where the transferred snapshot wins. GEAKG improves over Random Search on all six pairs.}
\label{fig:nas_heatmap_201}
\end{figure}

\begin{figure}[pos=!htbp]
\centering
\includegraphics[width=\columnwidth]{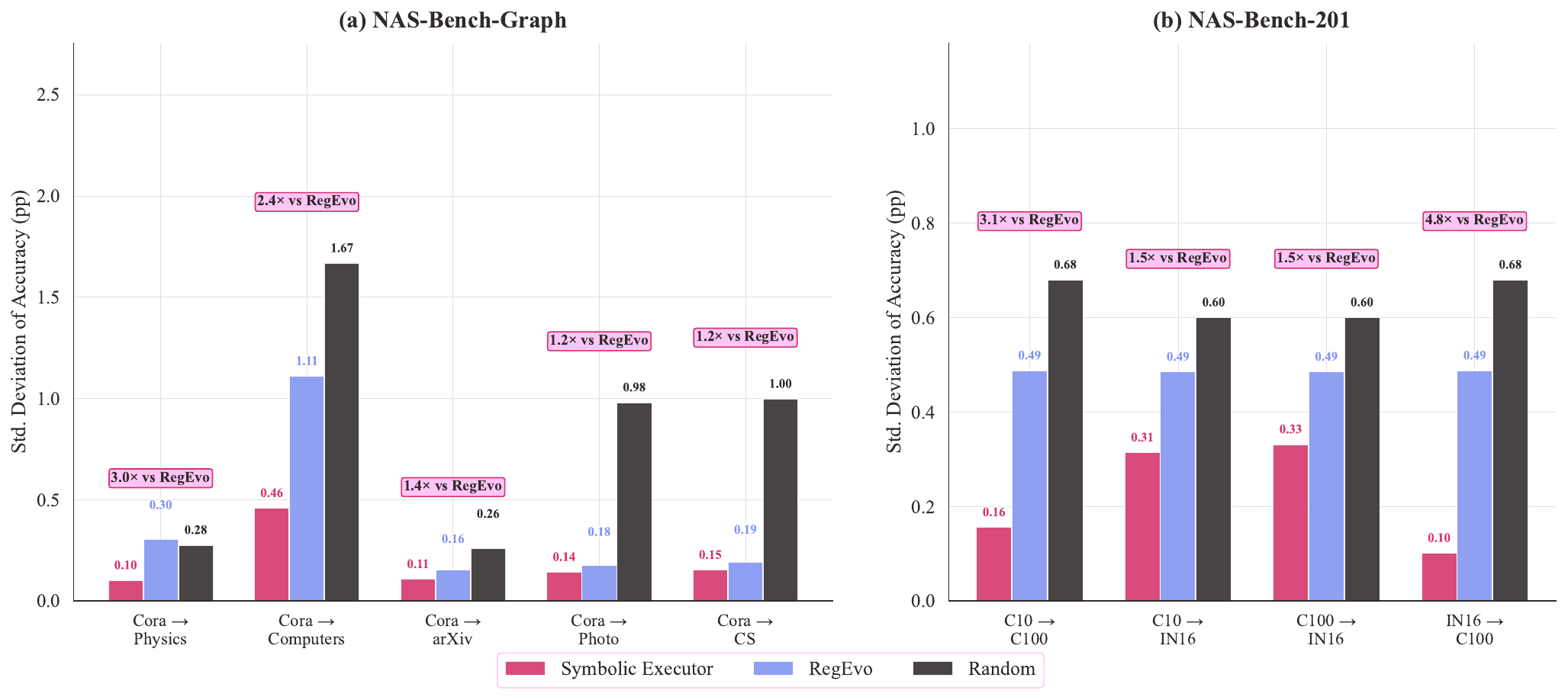}
\caption{Run-to-run variance of the GEAKG Symbolic Executor versus Regularized Evolution across representative NAS transfer pairs. Markers show each method's standard deviation of test accuracy over repeated runs. The Symbolic Executor has 1.2--4.8$\times$ lower standard deviation at comparable mean accuracy. This shows that a fixed procedural snapshot gives more repeatable behaviour than per-target evolutionary search.}
\label{fig:nas_variance}
\end{figure}

\begin{figure}[pos=!htbp]
\centering
\includegraphics[width=\columnwidth]{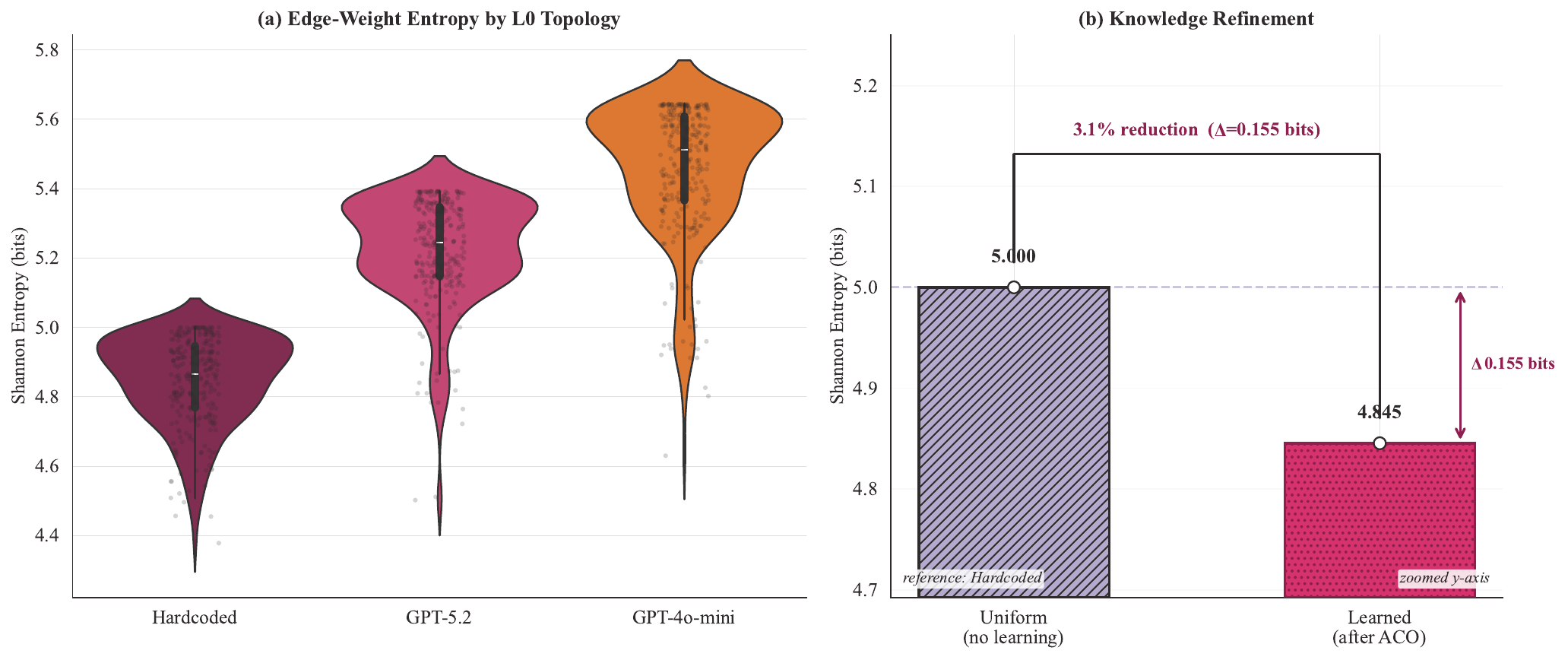}
\caption{Entropy of the learned edge-weight distribution in the NAS GEAKG. Results are grouped by topology source (baseline RoleSchema vs.\ LLM-generated topologies) and compare learned edge weights with a uniform baseline. Lower learned entropy means that ACO concentrates probability mass on preferred role transitions. MMAS bounds prevent collapse (a 3--4\% reduction from the uniform maximum).}
\label{fig:kg_pheromone_entropy}
\end{figure}

\begin{figure}[pos=!htbp]
\centering
\includegraphics[width=\columnwidth]{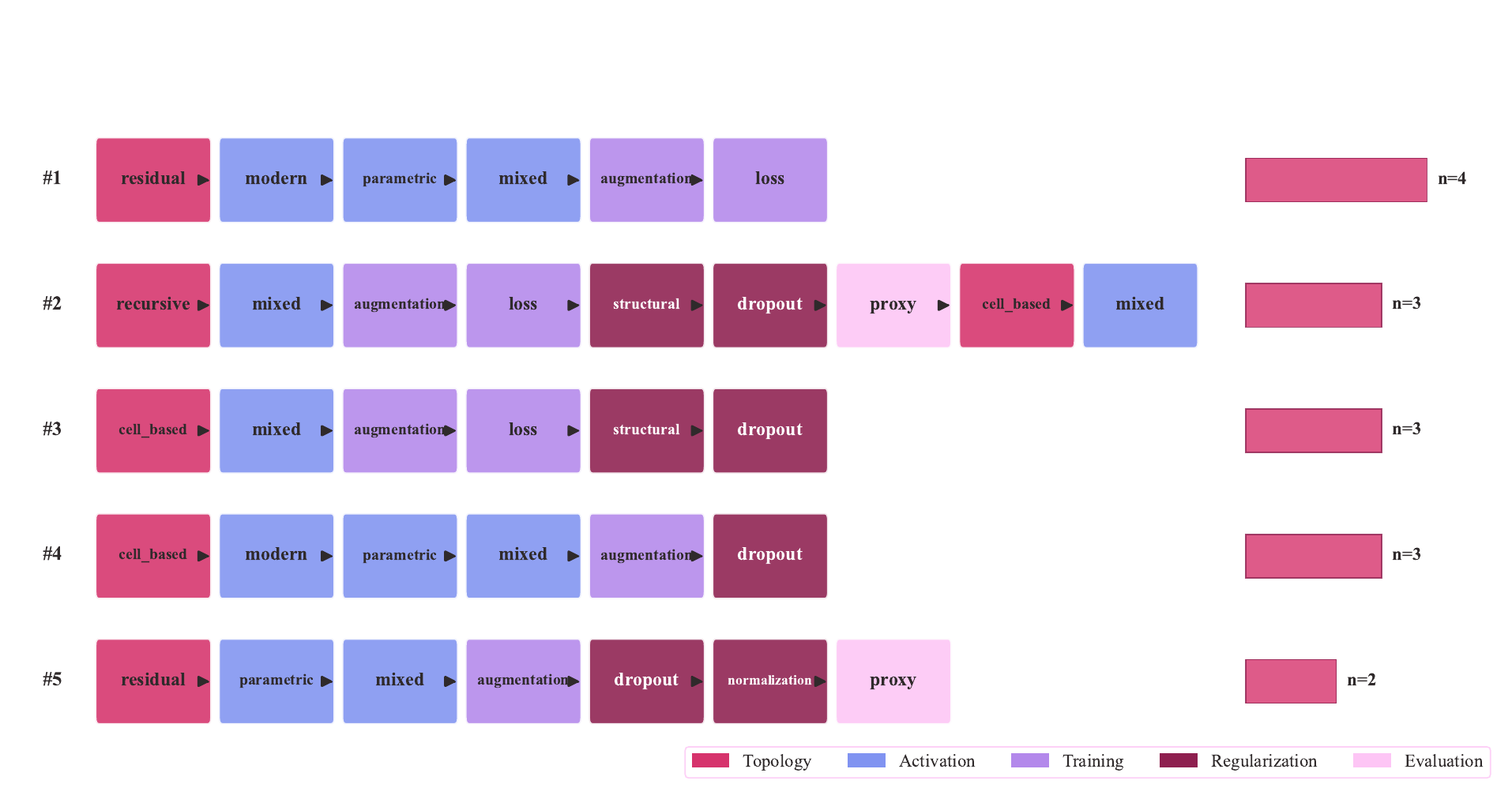}
\caption{Dominant traversal paths in the NAS GEAKG on NAS-Bench-Graph (Cora). The most frequent role sequences from top execution traces follow the category order Topology $\to$ Activation $\to$ Training $\to$ Regularization. Edge weight indicates how often each transition recurs. These paths are executable procedural sequences that the Symbolic Executor reuses at deployment.}
\label{fig:kg_dominant_paths}
\end{figure}

Supplementary Figures~\ref{fig:bifurcation}--\ref{fig:kg_dominant_paths} characterize the learned GEAKG artifact and its stability. They complement the artifact analysis in Methods. Supplementary Figure~\ref{fig:bifurcation} gives the perturbation (bifurcation) stability test. Supplementary Figures~\ref{fig:nas_heatmap_graph} and~\ref{fig:nas_heatmap_201} give the per-pair transfer gains over Random Search on the two NAS benchmarks. Supplementary Figure~\ref{fig:nas_variance} reports the variance advantage over Regularized Evolution. Supplementary Figures~\ref{fig:kg_pheromone_entropy} and~\ref{fig:kg_dominant_paths} show the learned edge-weight-entropy concentration and the dominant traversal paths through the role graph.

\subsection*{Supplementary Figures}

\end{document}